\documentclass{IEEEtran}

\usepackage{ifluatex,ifxetex}
\ifxetex
  \usepackage{fontspec}
\else
  \ifluatex
    \usepackage{fontspec}
  \else
    \usepackage[utf8]{inputenc}
  \fi
\fi

\usepackage{bm}

\usepackage{breqn}
\usepackage{subfig}

\usepackage{pifont}
\usepackage{rotating}
\usepackage{afterpage}
\usepackage{array}
\usepackage{amsfonts}
\usepackage{amsmath}
\usepackage{amssymb}
\usepackage{cite}
\usepackage{color}
\usepackage{enumerate}
\usepackage{epsfig}
\usepackage{graphicx}
\usepackage{multirow}
\usepackage{setspace}
\usepackage{stmaryrd}
\usepackage{booktabs}
\usepackage[usenames,dvipsnames]{xcolor}
\usepackage{algorithmic}  
\usepackage{algorithm}

\newtheorem{theorem}{Theorem}

\makeatletter
\usepackage{multirow,booktabs,color,soul,threeparttable}
\definecolor{hl}{rgb}{0.75,0.75,0.75}
\sethlcolor{hl}

\usepackage{float}\@ifundefined{definecolor}
{\usepackage{color}}{}
\makeatother

\begin{document}
\def\ie{\emph{i.e.}}
\def\eg{\emph{e.g.}}%
\def\et{\emph{et al.}}
\newcommand{\zk}[1]{{\color{magenta}{[zk:#1]}}}
\newcommand{\Res}[1]{{\color{BlueGreen}{[Li:#1]}}}

\title{Analyzing and Overcoming Local Optima in Complex Multi-Objective Optimization by Decomposition-Based Evolutionary Algorithms}  

\author{Ting Dong, Haoxin Wang, Hengxi Zhang, Wenbo Ding}

\maketitle
\begin{abstract}

When addressing the challenge of complex multi-objective optimization problems, particularly those with non-convex and non-uniform Pareto fronts, Decomposition-based Multi-Objective Evolutionary Algorithms (MOEADs) often converge to local optima, thereby limiting solution diversity. Despite its significance, this issue has received limited theoretical exploration. Through a comprehensive geometric analysis, we identify that the traditional method of Reference Point (RP) selection fundamentally contributes to this challenge. In response, we introduce an innovative RP selection strategy, the Weight Vector-Guided and Gaussian-Hybrid method, designed to overcome the local optima issue. This approach employs a novel RP type that aligns with weight vector directions and integrates a Gaussian distribution to combine three distinct RP categories. Our research comprises two main experimental components: an ablation study involving 14 algorithms within the MOEADs framework, spanning from 2014 to 2022, to validate our theoretical framework, and a series of empirical tests to evaluate the effectiveness of our proposed method against both traditional and cutting-edge alternatives. Results demonstrate that our method achieves remarkable improvements in both population diversity and convergence.



\end{abstract}

\section{Introduction}
\label{sec:intro}


\IEEEPARstart{M}{ulti-objective} optimization problems (MOPs) are referred to those required to optimize more than two conflicting objectives at the same time. Pareto front (PF) of a MOP is a set of mapping of non-dominated solutions, representing the best trade-offs in the objective function space. Solutions on the PF are considered optimal because improving one objective would lead to the deterioration of at least one other objective. The decomposition-based multi-objective evolutionary algorithm  (MOEA/D)~\cite{MOEAD} has become one popular framework to solve MOPs since it was presented. Within this framework, many variants of MOEA/D (MOEA/D-variants) have been developed~\cite{CMOEAD,MOEADAWA,MOEADM2M,MOEADMRDL,BCEMOEAD,MOEADDU,MOEADPas,MOEADDCWV,MOEADURAW,MOEADDYTS,MOEADUR,MOEADVOV,repoint1,repoint2,otherMethod,weight1,decompositionMethods1,decompositonmethods2} over the past decade, focusing on enhancing different components of the algorithm. A common phenomenon is that when dealing with complex MOPs, especially those characterized by non-convex and non-uniform PFs, almost all MOEA/D-variants rapidly lose their effectiveness in maintaining diversity~\cite{irregularDiversity1,irregularDiversity2,irregularDiversity3,irregularDiversity4,irregularDiversity5}. This means that the population has fallen into local optima during the iteration process. 

As Hisao et al. proposed~\cite{performanceMOEAD}, the performance of MOEAD-variants algorithms strongly depends on the shapes of the PF. Specifically, a slight change in the problem formulations of DTLZ and WFG can have a detrimental impact on the performance of  MOEA/D-based Algorithms. When the PF is convex, MOEA/D algorithms still achieve satisfactory results, including both population diversity and convergence. However, when the PF is non-convex, especially with highly concave shapes such as in the IMOP and DTLZ2 problems, the performance of  suffers a noticeable decline in both convergence and diversity.

Although this phenomenon is widespread, there is no rigorous theoretical exploration to answer why almost all MOEA/D-variants exhibit this phenomenon. At present, most studies are based on experimental simulations followed by qualitative analysis to explain the reason for the existence of question~\cite{irregularDiversity1,irregularDiversity2,irregularDiversity3,irregularDiversity4,irregularDiversity5,performanceMOEAD}.In this paper, we provide our rigorous geometric analysis for this critical issue and introduce a novel theoretically supported method to address this problem. specifically, we find under the conditions where weight vectors, genetic operators, and decomposition methods all adopt their default values, the traditional method of selecting the reference point, the default \textit{min} method, stands out as the primary reason for the decline in algorithmic performance. This work intends to make the following contributions.

\begin{enumerate}
    \item It proposes a geometric proof demonstrating that the traditional method of selecting the RP is the primary reason why MOEA/D-variants tend to fall into local optima in complex MOPs. Furthermore, it empirically validates the accuracy of this theory, through the ablation study~\cite{ablation} on selected fourteen MOEA/D-variants  published between 2014 and 2022.
    
    \item   It proposes a efficient and novel strategy to select the RP, the Weight Vector-Guided and Gaussian-Hybrid method. The method generates a new type of RP along the direction of the weight vectors to ensure diversity and employs Gaussian distribution to ensure convergence.  
   
    
    
    \item The proposed strategy outperforms not only the traditional but also the state-of-the-art method in terms of diversity, while reaching the same level of them in convergence.
\end{enumerate}

The remainder of this article is organized as follows. Section \ref{sec:backg} introduces relevant definitions and existing methods. Section \ref{sec:geometricProof} provides a geometric analysis of why the MOEA/D framework tends to converge to local optima in complex MOPs. Following this, Section \ref{Methodology} details the methodology of the proposed method. Section \ref{experiments} presents an ablation study along with experiments on the proposed method. Finally, Section \ref{conclusion} concludes the article.



\section{Background}
\label{sec:backg}
\subsection{MOPs}
Multi-objective optimization problems (MOPs) are formulated as follows:
\begin{equation}
\begin{array}{ll}
\text{minimize} & \mathbf{F}(\mathbf{x}) = (f_{1}(\mathbf{x}), \ldots, f_{M}(\mathbf{x}))^\intercal, \\
\text{subject to} & \mathbf{x} \in \Omega \subset \mathbb{R}^D,
\end{array}
\label{MOP}
\end{equation}
where $\Omega \subset \mathbb{R}^{D}$ represents the decision space in $D$ dimensions. The decision vector $\mathbf{x} = (x_{1}, \ldots, x_{D})^\intercal$ and its associated objective vector $\mathbf{F}(\mathbf{x})$ are central to the optimization. The functions $f_{1}(\mathbf{x}), \ldots, f_{M}(\mathbf{x})$ serve to define the mapping from the decision space to the objective space.

\subsubsection{Pareto-Dominate}Given two solutions $\mathbf{x}_{1}$ and $\mathbf{x}_{2}$, we say $\mathbf{x}_{1}$ Pareto-dominates $\mathbf{x}_2$ (expressed as $\mathbf{x}_{1} \preccurlyeq \mathbf{x}_{2}$) if $f_{i}(\mathbf{x}_{1}) \leq f_{i}(\mathbf{x}_{2})$ across all objectives $i \in \{1, \ldots, M\}$, and there exists at least one objective $j \in \{1, \ldots, M\}$ where $f_{j}(\mathbf{x}_{1}) < f_{j}(\mathbf{x}_{2})$. 
\subsubsection{Pareto Front} A collection of optimal compromise solutions, termed the Pareto set ($\mathbf{PS}$), defined as $\mathbf{PS} = \{\mathbf{x}^{*} \in \Omega \ | \ \nexists \mathbf{x} \in \Omega, \mathbf{x} \preccurlyeq \mathbf{x}^{*}\}$. The mapping of $\mathbf{PS}$ within the objective space is known as the Pareto front ($\mathbf{PF}$), represented by $\mathbf{PF} = \{\mathbf{f}(\mathbf{x}^{*}) \ | \ \mathbf{x}^{*} \in \mathbf{PS}\}$.

The aim of multi-objective  optimization is to obtain a group of non-dominated solutions as a result of the trade-off between different objectives. Therefore, an algorithm must guarantee both diversity and convergence. Under ideal conditions, the mapping of candidate solutions can be evenly distributed on the real PF.

\subsection{MOEA/D}
\label{definition}

By utilizing weight vectors, MOPs are decomposed into several subproblems, with the goal of each subproblem being to minimize a specific weighted objective function. An initial population \( P \) and its corresponding neighborhood \( T \) are established. Subsequently, offspring \( y \) are produced via crossover and mutation techniques, with their performance evaluated using a fitness function. If \( y \) demonstrate superior performance compared to the neighboring parents, it then takes their place. Below is the pseudocode \ref{alg:MOEAD} for MOEA/D outlined.

\begin{figure}[h]
\begin{algorithm}[H]
\caption{MOEA/D }
\label{alg:MOEAD}
\begin{algorithmic}[1]

\STATE \textbf{Input:} 
\begin{itemize}
    \item Population size $N$
    \item the  number of function evaluations $maxFE$
    \item Number of objectives $M$
    \item Crossover and mutation rates
    \item Neighborhood size $T$
\end{itemize}
\STATE \textbf{Output:} A population set P

\STATE Initialize population $P$ of size $N$;
\STATE Evaluate each solution in $P$;
\STATE Initialize weight vectors  $W_1, W_2,...,W_N$ for each solution;
\STATE Calculate Euclidean distances between weight vectors, and then determine the $T$ weight vectors closest to each weight vector;
\STATE These form the neighborhoods of $i, i=1,2,...,N$;
\STATE Initialize the  reference point $Z = (z_i^{min}),i=1,2,...,M$,where $z_{i}^{min}$ is the minimum  values of the $i$th objective of all solutions in objective space;

\FOR{$FE = 1$ to $maxFE$}
    \FOR{each solution $i$ in $P$}
        \STATE Select two parents $p_1$ and $p_2$ from the neighborhood of $i$
        \STATE Generate an offspring $y$ by applying crossover and mutation operators to $p_1$ and $p_2$
        \STATE Generate the $Z = min(Z,y)$
        \STATE Evaluate $y$ by the aggregation of objectives using the weight vector of $i $ $(W_i)$ and $Z$

        \FOR{each solution $j$ in the neighborhood of $i$}
            \IF{$y$ is better than the $j$-th solution with respect to their own decomposition function $g(W_i,Z)$}
                \STATE Replace the $j$-th solution with $y$ in P
            \ENDIF
        \ENDFOR
    \ENDFOR
\ENDFOR
\STATE \textbf{return}  $P$

\end{algorithmic}
\end{algorithm}
\end{figure}

\subsubsection{\textbf{Reference point}}
In MOEA/D-based algorithms, the reference point~\cite{MOEAD} is utilized to guide the algorithm's search direction and aid in a more accurate approximation to the PF. It gets updated throughout the algorithm's execution.

\subsubsection{\textbf{True ideal point}}
For a multi-objective minimization problem, each component of the true ideal point represents the minimum value of its corresponding objective function across the entire feasible solution space. However, in practical optimization processes, the true ideal point is unknown. The aforementioned reference point is an obtained ideal point maintained and computed by the algorithm using certain methods.





\subsubsection{\textbf{Fitness function}}




A fitness function \cite{fitness} is an evaluation function used to assess and compare the quality of candidate solutions in an optimization problem. In MOPs, One common strategy is to use  the decomposition methods as fitness functions. Another common strategy is to use Pareto ranking(PR)~\cite{NSGAII}.


\subsubsection{\textbf{Decomposition method}}


In some literature, decomposition methods are also referred to as scalar methods. They are a category of fitness functions in MOPs used to assess and compare the quality of candidate solutions. This includes Weighted Sum Approach (WS)~\cite{WS}, Tchebycheff Approach (TCH)~\cite{WS}, the penalty-based boundary intersection approach (PBI)~\cite{MOEAD}, among others. In the \textit{Geometric Proof} section of this paper, we employed PBI, while in the \textit{Experiments} phase, we utilized Modified Tchebycheff Approach (M-TCH)~\cite{m-TCH}. The formulas for these methods are presented below. Let Equation\ref{eq:pbi} correspond to PBI and Equation\ref{eq:mtch} correspond to M-TCH.

\begin{equation}
    \begin{aligned}\min_{\mathbf{x}\in\Omega}&g^{pbi}(\mathbf{x}|\mathbf{W},\mathbf{Z})=d_1+\theta d_2\\
    d_1&=\frac{\|(\mathbf{F}(\mathbf{x}) - \mathbf{Z})^T\mathbf{W}\|}{\|\mathbf{W}\|}\\
    d_2&=\|\mathbf{F}(\mathbf{x})-(\mathbf{Z}+d_1\mathbf{W})\|.\end{aligned}
    \label{eq:pbi} 
\end{equation}

\begin{equation}
\min_{\mathbf{x}\in\Omega}g^{\mathrm{mtch}}\big(\mathbf{F}(\mathbf{x})|\mathbf{W},\mathbf{Z}\big)=\max_{1\leq i\leq m}\bigg\{\frac{f_i(\mathbf{x})-z_i}{W_i}\bigg\}.
\label{eq:mtch} 
\end{equation}
where \( X \) represents the decision vector, \( F(X) \) denotes the objective vector, \( W \) is the weight vector, and \( Z \) is the reference point. \( f_i(x) \) signifies the objective vector of the \( i^{th} \) subproblem, \( z_i \) represents the reference point corresponding to the \( i^{th} \) subproblem, and \( W_i \) is the weight vector associated with the \( i^{th} \) subproblem.

\subsection{Existing Methods}\label{sec:existingmethod}

We will begin by discussing the limitations of existing methods, introduce two prevalent methods, and then delve into the method we propose.
In existing methods, as evidenced from the proof section, the traditional min method tends to lead the algorithm straight into local solutions for concave problems. An issue with the more recent DRP method is that, since the reference point in DRP has a linear relationship with FE, the HV metric that measures the diversity of the population and the IGD metric that gauges convergence still exhibit considerable fluctuations, even after the algorithm has converged. The metrics even display a declining trend. Essentially, the DRP method\cite{otherMethod} (as seen in Equation \ref{eq:DRP}) is an improvement upon the min method, but it doesn't fully address the challenge of the algorithm getting stuck in local solutions. These conclusions can be verified in the experimental section. Below are the formulas for both methods.
\subsubsection{\textbf{The Traditional Method}}
With the introduction of MOEA/D\cite{MOEAD}, the selection method of the reference point, commonly known as the min method, has been extensively utilized in various enhanced algorithms. Its formula is as follows.
\begin{align}
\begin{aligned}
 Z_m &=(z_{1}^{min},\ldots,z_{m}^{min})^T,\\ 
    z_{i}^{min} &= \min\{f_i(x)|x\in\Omega\}, \text{ for each }i=1,\ldots,m.
\end{aligned}
\label{eq:zmin} 
\end{align}

\subsubsection{\textbf{The Latest Method}}
In this paper\cite{otherMethod}, the \textit{dynamic reference point (DRP)} method is introduced . By incorporating the variable \( \epsilon_i \) into \( z_i^{min} \), it derives the following formula.
\begin{align}
\begin{aligned}
Z_d &= (z_{1}^{drp},\ldots,z_{m}^{drp})^T,\\
z_i^{drp}&=z_i^{\min}-\epsilon_i,\\\epsilon_i &=(\epsilon_i^{ini}-\epsilon_i^{end})\left(\frac{FE- maxFE }{FE-1}\right)+\epsilon_i^{end}, \\\quad\epsilon_i^{ini}&=1,\quad\epsilon_i^{end}=0.001.
\end{aligned}
\label{eq:DRP} 
\end{align}

\section{Geometric Proof} \label{sec:geometricProof}
This chapter aims to perform a theoretical analysis on how MOEA/D escapes LO in IMOP2. A significant challenge currently faced by decomposition-based multi-objective evolutionary algorithms is their propensity to fall into LO when MOEA/D-variants address MOPs characterized by non-convexity and non-uniformity of the PF \cite{MOEAD}. To derive universally applicable conclusions, this chapter selects the classic MOEA/D and the classical problem IMOP2, which exhibits both characteristics, for theoretical analysis. It proves that the fundamental reason for MOEA/D's frequent entrapment in LO on complex MOPs is the traditional RF selection method: min. The first Section~\ref{preliminaries} introduces the properties of the IMOP2 problem, followed by sections~\ref{resultsPBI} and~\ref{resultsMTCH}, which analyze why MOEA/D cannot escape LO under the traditional min method.
\subsection{Preliminaries}
\label{preliminaries}
The representative IMOP2 problem has decision variables of dimension \(K+L\) and two objective functions as follows:
\begin{equation}
\text{min}\quad
\begin{aligned}
f_1(\bm{x})=\sqrt{\text{cos} \left(\frac{2}{\pi}y\right)}+g,\\
f_2(\bm{x})=\sqrt{\text{sin} \left(\frac{2}{\pi}y\right)}+g,
\end{aligned}
\end{equation}
where
\begin{equation}
\begin{aligned}
y=\left(\frac{1}{K} \sum _{i=1} ^K x_i\right)^{a},\ a = 0.05,\\
g=\sum _{i=K+1} ^{K+L} \left(x_i-\frac{1}{2}\right)^2,\\
x_1,...,x_{K+L} \in [0,1].
\end{aligned}
\end{equation}

Notice that when \(x_{K+1}=...=x_{K+L}=\frac{1}{2}\), \(g=0\), and for all \(x_1,...,x_K \in [0,1], y \in [0,1]\), thus by eliminating the parameter \(y\), the Pareto Front of IMOP2 can be expressed as:
\begin{equation}
f_2=(1-f_1^4)^{\frac{1}{4}},\ 0 \leq f_1 \leq 1.
\end{equation}

\textbf{Nonconvexity} is simply determined by calculating \(\frac{\text{d}^2 f_2}{\text{d} f_1^2} = -3f_1^2 (1-f_1^4)^{-\frac{7}{4}} < 0\), hence the PF is a concave function. Furthermore, we calculate the area of the curvilinear triangle enclosed by the PF and the \(f_1, f_2\) axes to be \(S= \int _0 ^1 (1-f_1^4)^{\frac{1}{4}} d f_1 = 2 \pi^{-\frac{1}{2}} \Gamma ^2(\frac{5}{4}) = 0.93 \approx 1\), indicating that the PF is highly concave. The image of the PF is shown in Figure~\ref{fig6.1}.

\textbf{Nonuniformity} results from \(g=\sum _{i=K+1} ^{K+L} (x_i-\frac{1}{2})^2\) being a quadratic function in terms of variables \(x_{K+1},...,x_{K+L}\), which can be easily optimized by algorithms, thus in most cases, it is easy for algorithms to find points on the PF. Based on this, since the value of \(a\) is very small (default value 0.05) \cite{IMOP}, \(y\) is close to 1 in most cases, \(\sqrt{\text{cos} \frac{\pi}{2}y}\) is close to 0, and \(\sqrt{\text{sin} \frac{\pi}{2}y}\) is close to 1, meaning the PF solutions are nonuniformly distributed along the global PF, and the PF is biased towards solutions for which \(f_1(\bm{x})=0\). 

\begin{figure}[htbp]
\centering
\subfloat[]{
\includegraphics[width=0.45\textwidth]{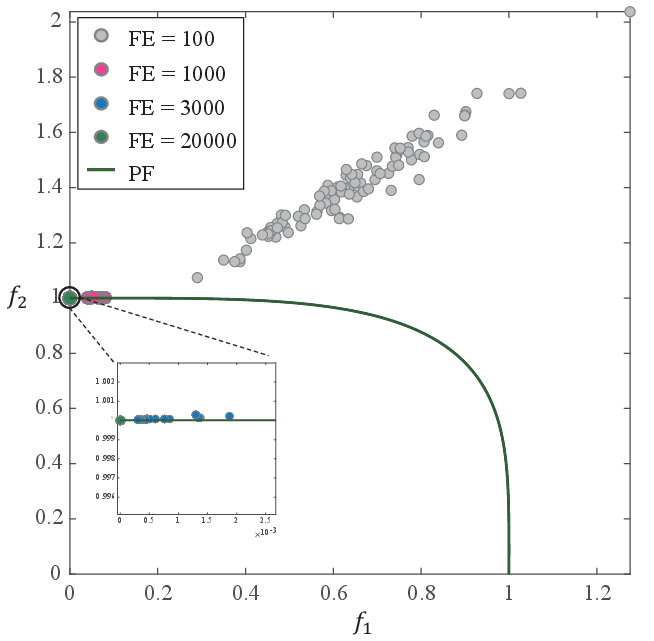}
\label{fg:IMOP2wlinear}}
\hfil
\subfloat[]{
\includegraphics[width=0.45\textwidth]{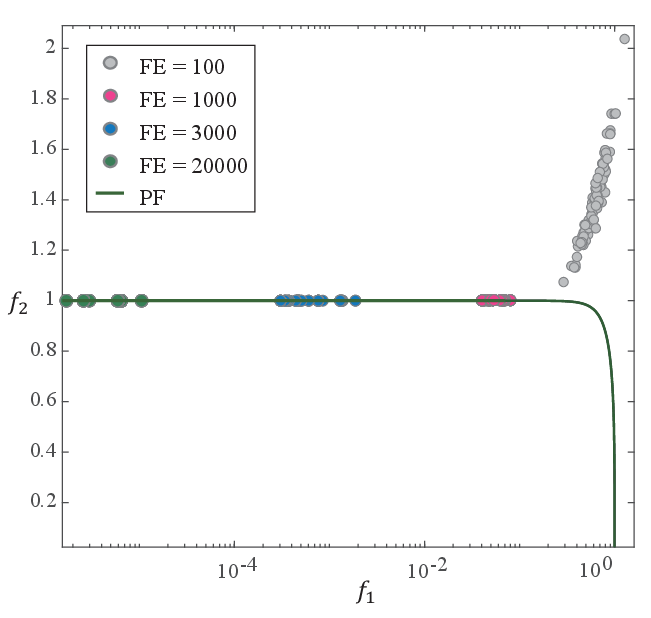}
\label{fg:IMOP2wlog}}
\caption{Illustration of the population results obtained by MOEA/D on the IMOP2 problem with $FE=20000$: (a) linear coordinate axis; (b) logarithmic coordinate axis.}
\label{fig6.2}
\end{figure}

\begin{figure}[htbp]
\centering
\subfloat[]{
\includegraphics[width=0.22\textwidth]{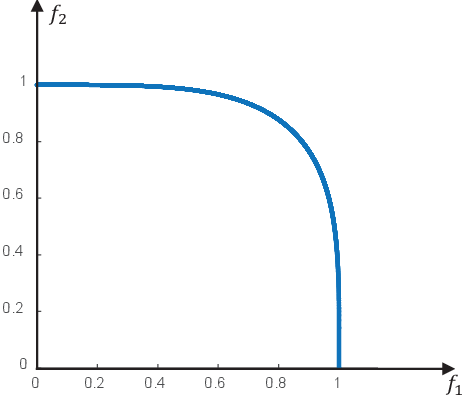}
\label{fig6.1}}
\subfloat[]{
\includegraphics[width=0.24\textwidth]{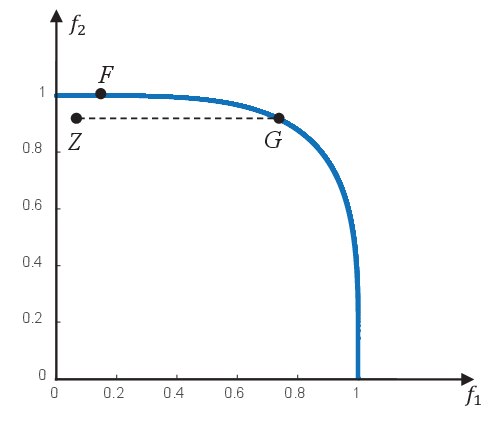}
\label{fig6.3}}
\caption{The Pareto Front of IMOP2 and Positions of \(F, Z, G\) under the objective function coordinate plane.}
\label{fig6.12}
\end{figure}

Due to the nonuniformity of IMOP2, it is likely that the population in the algorithm will not be uniformly distributed on the PF but will stagnate near \( (f_1,f_2)=(0,1)\). To illustrate this, we solve IMOP2 using the MOEA/D algorithm and observe the distribution of the population at \(FE=10,1000,3000,20000\) as shown in Figure \ref{fig6.2}. As seen in Figure \ref{fig6.2}, with the increase in the number of iterations, MOEA/D gradually stagnates near \( (f_1,f_2)=(1,0)\). The points with different colors represent the parent populations \(P\) obtained at different evolutionary stages. The first graph uses a linear axis, while the second employs a logarithmic axis, providing a clearer representation of the obtained population set \(P\).

Combining the above discussions, in the following analysis, we assume that MOEA/D has stagnated, that is, in the objective space, all individual positions are located near \( (f_1,f_2)=(0,1)\). We assume the position of a certain individual as \(\bm{x}=[x_1,...,x_{K+L}]\), whose coordinates in the objective function coordinate plane are \(F=(F_1,F_2)\), considering that the \(g\) part can be easily optimized, so we can set \(x_{K+1}=...=x_{K+L}=0.5\), i.e., \(F\) is on the Pareto Front, and close to \((1,0)\). Clearly, according to the definition of the min method, at this time, the algorithm's Reference Point (RP) \(Z=[Z_1,Z_2]\) is located below and to the left of \(\bm{F}\), i.e., \(Z_1 \leq F_1, Z_2 \leq F_2\).

Based on this, we assume the position of a newly generated individual as \(\bm{y}=[y_1,...,y_{K+L}]\), with its coordinates in the objective function space being \(G=(G_1,G_2)\). Since all progenitor positions from the \(K+1\)th to \(K+L\)th components are approximately 0.5, we can set \(y_{K+1}=...=y_{K+L}=0.5\), meaning \(G\) is on the Pareto Front. Further, we assume that \(G\) is to the right of \(F\), i.e., \(G_1>F_1,G_2<F_2\). This assumption has clear significance: if the population generates the individual \(G\) in some iteration, and if the algorithm can accept this individual with a relatively high probability, then the population can gradually move to the right, thereby ensuring the algorithm's ability to escape LO; if the algorithm can hardly accept this individual, it is considered difficult for the algorithm to escape LO. Further, after generating individual \(G\), the position of RP updates to \(Z=[Z_1,G_2]\). Assuming the distance between \(G\) and \(F\) is not too large (offspring generated from parents naturally have a higher probability of being near the parents), and the value of \(G_1\) is closer to 0, we assume the slope of the PF at \(G\) as \(k_G\), then at this time
\begin{equation}
|k_G| = \frac{G_1^3}{(1-G_1^4)^{\frac{3}{4}}}
\end{equation}
is evidently very small.

Finally, we assume the positions of \(F, Z\), and \(G\) under the objective function coordinate plane as shown in Figure~\ref{fig6.3}, here \(GZ\) is parallel to the \(f_1\) axis.


\subsection{Results under PBI}
\label{resultsPBI}
For MOEA/D employing the PBI method, this section will analyze the possibility of the algorithm accepting a new individual \(G\) based on Figure~\ref{fig6.3}. Specifically, given the point \(W=[W_1,W_2]\), we only need to compare the magnitude of \(g^{pbi}(F|W,Z)\) and \(g^{pbi}(G|W,Z)\), where \(g^{pbi}\) is the fitness function in the PBI method, with its expression seen in a certain formula~\ref{eq:pbi}. The analysis unfolds under two scenarios: (1) \(k_{OW} > k_{ZF}\), and (2) \(k_{OW} \leq k_{ZF}\), where \(k_{XY}\) represents the slope of the line \(XY\), defined as \(+\infty\) when \(XY\) is parallel to the \(f_2\) axis, and similarly below. Moreover, we define the angle between \(\overrightarrow{OW}\) and \(Of_1\) as \(\alpha\).

First, we discuss the scenario where \(k_{OW} > k_{ZF}\). We present and prove the following theorem:
\begin{theorem}
\label{theorem1}
Given \(k_{OW} > k_{ZF}\), if
\begin{equation}
|k_G| \leq \theta
\end{equation}
then
\begin{equation}
g^{pbi}(F|W,Z) < g^{pbi}(G|W,Z)
\end{equation}
\end{theorem}

\begin{IEEEproof}
The objective function coordinate plane for this scenario is depicted in Figure~\ref{fig6.4}. Here, \(AD \perp OW\), \(AG \perp Of_1\), \(AF \parallel ZG \parallel Of_1\), \(BF \parallel CG \parallel AD\), \(AH \parallel FE \parallel OW\) are auxiliary lines. By the geometric meaning of PBI,
\begin{equation*}
g^{pbi}(F|W,Z)=|ZB|+\theta |BF|
\end{equation*}
\begin{equation*}
\begin{aligned}
g^{pbi}(G|W,Z)=&|ZC|+\theta |CG|\\
=&|ZB|+|BD|-|CD|+ \theta (|CE|+|EH|+|HG|)\\
=&|ZB|+|AF|\cos\alpha -|AG|\sin\alpha \\
+& \theta(|BF|+|AF|\sin\alpha+|AG|\cos\alpha)
\end{aligned}
\end{equation*}

Thus,
\begin{equation*}
\begin{aligned}
g^{pbi}(G|W,Z)-&g^{pbi}(F|W,Z) \\
=& |AF|\cos\alpha -|AG|\sin\alpha \\
+& \theta(|AF|\sin\alpha+|AG|\cos\alpha)  \\
=& (|AF|+\theta|AG|)\cos\alpha + (\theta|AF|-|AG|)\sin\alpha\\
>& (\theta|AF|-|AG|)\sin\alpha\\
=& |AF|(\theta-|k_{FG}|)\sin\alpha\\
>& |AF|(\theta-|k_{G}|)\sin\alpha \geq 0
\end{aligned}
\end{equation*}

\begin{figure}[htbp]
  \centering
  \includegraphics[width=0.8\linewidth]{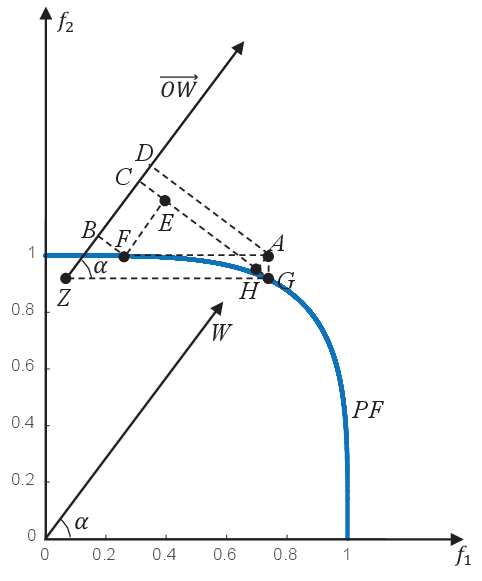}
  \caption{Objective function coordinate plane under the condition \(k_{OW} > k_{ZF}\)}
    \label{fig6.4}
\end{figure}

\end{IEEEproof}
Next, we discuss the scenario where \(k_{OW} \leq k_{ZF}\). We propose and prove the following theorem:
\begin{theorem}
\label{theorem2}
Given \(k_{OW} \leq k_{ZF}\), if
\begin{equation}
|k_G| \leq \min \left\{ \frac{1}{2} \sin^3 \alpha \cos \alpha , \frac{1}{2} \cot \alpha \right\}
\end{equation}
then
\begin{equation}
g^{pbi}(F|W,Z) < g^{pbi}(G|W,Z)
\end{equation}
\end{theorem}

\begin{IEEEproof}
The objective function's coordinate plane under this condition is illustrated in Figure~\ref{fig6.5}. Here, \(AC\) is tangent to the Pareto Front at point \(A\), \(BG \perp OW\), and \(DE \perp OW\).

First, we estimate the upper bound of \(g^{pbi}(F|W,Z)\) as follows:
\begin{equation*}
\begin{aligned}
g^{pbi}(F|W,Z) &= |ZD|+\theta |DF| \\
&< |ZD|+\theta |ED|\\
&= |AZ|-|AD|+\theta |AD| \tan \angle CAZ \\
&= |AZ| + |AD| \left(\theta \frac{ |k_{AC}|+\tan \alpha }{1 -|k_{AC}| \tan \alpha  }-1\right) \\
&\leq |AZ| \max \left\{ 1,\theta \frac{ |k_{G}|+\tan \alpha }{1 -|k_{G}| \tan \alpha  } \right\}\\
&\leq |AZ|  \left(\cot \alpha + 2\tan \alpha\right)\theta \\
&< |AZ|\frac{2\theta}{\sin \alpha \cos \alpha}
\end{aligned}
\end{equation*}

Next, we estimate the lower bound of \(g^{pbi}(G|W,Z)\). Obviously,
\begin{equation*}
|AG|=\sqrt{|AB|^2+|BG|^2}=\sqrt{|AB|^2+\tan^2 \alpha (|AZ|+|AB|)^2}
\end{equation*}

Let \(\angle AGZ =\beta\). Using the sine theorem in \(\triangle AZG\):
\begin{equation*}
\frac{|AG|}{\sin \alpha} = \frac{|AZ|}{\sin \beta}
\end{equation*}

Combining the above equations, we get
\begin{equation*}
|AB|=|AZ|(\sin \alpha \cos \alpha \cot \beta - \sin^2 \alpha)
\end{equation*}

Therefore,
\begin{equation*}
\begin{aligned}
&g^{pbi}(G|W,Z) = |BZ| + \theta |BG| \\
&= (1+\theta \tan \alpha) (|AZ|+|AB|) \\
&= |AZ|(1+\theta \tan \alpha)(\sin \alpha \cos \alpha \cot \beta +\cos^2 \alpha)\\
&> |AZ|\theta \frac{\sin^2 \alpha}{|k_G|}\\
&\geq |AZ|\frac{2\theta}{\sin \alpha \cos \alpha}\\
&> g^{pbi}(F|W,Z)
\end{aligned}
\end{equation*}
\begin{figure}[htbp]
  \centering
  \includegraphics[width=\linewidth]{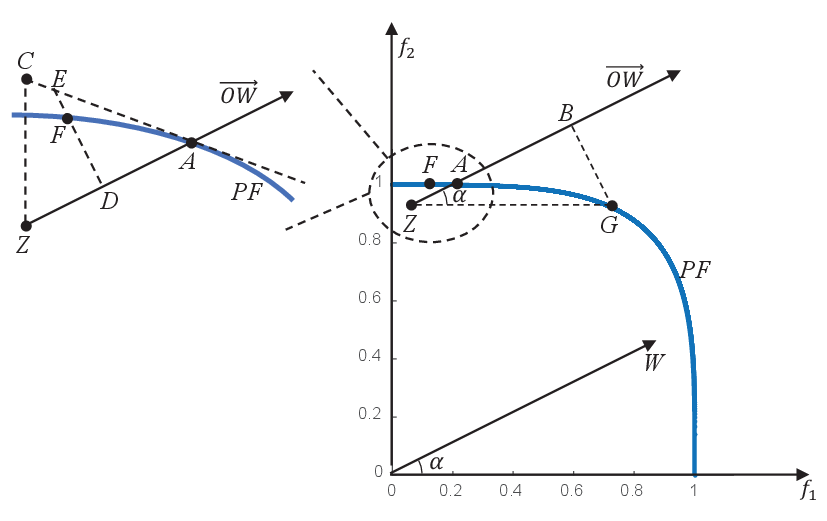}
  \caption{The objective function coordinate plane under the condition \(k_{OW} \leq k_{ZF}\)}
    \label{fig6.5}
\end{figure}

\end{IEEEproof}

In summary, revisiting Theorems~\ref{theorem1} and~\ref{theorem2}, as long as the respective assumptions are satisfied, \(g^{pbi}(F|W,Z) < g^{pbi}(G|W,Z)\) holds true, meaning the fitness of individual \(G\) is worse than that of \(F\), implying the new individual \(G\) cannot be accepted, and the algorithm struggles to escape local optima. As discussed in the Preliminaries chapter, the value of \(|k_G|\) is generally very small, thus for most weight vectors \(W\), the respective assumptions are easily satisfied. In conclusion, the probability of the algorithm escaping LO is very small.
\subsection{Results under M-TCH}
\label{resultsMTCH}
For MOEA/D employing the M-TCH method, this chapter will analyze the possibility of the algorithm accepting a new individual \(G\) based on Figure~\ref{fig6.3}. Specifically, given the point \(W=[W_1,W_2]\), we only need to compare the magnitude of \(g^{mtch}(F|W,Z)\) and \(g^{mtch}(G|W,Z)\), where \(g^{mtch}\) is the fitness indicator within the M-TCH method, with its expression found in a certain formula~\ref{eq:mtch}. The analysis will be conducted under two scenarios: (1) \(k_{OW} > k_{ZF}\), and (2) \(k_{OW} \leq k_{ZF}\). Additionally, we define the angle between \(\overrightarrow{OW}\) and \(Of_1\) as \(\alpha\).

First, we discuss the scenario where \(k_{OW} > k_{ZF}\). We propose and prove the following theorem:
\begin{theorem}
\label{theorem3}
Given \(k_{OW} > k_{ZF}\), then
\begin{equation}
g^{mtch}(F|W,Z) < g^{mtch}(G|W,Z)
\end{equation}
\end{theorem}

\begin{figure}[htbp]
  \centering
  \includegraphics[width=0.8\linewidth]{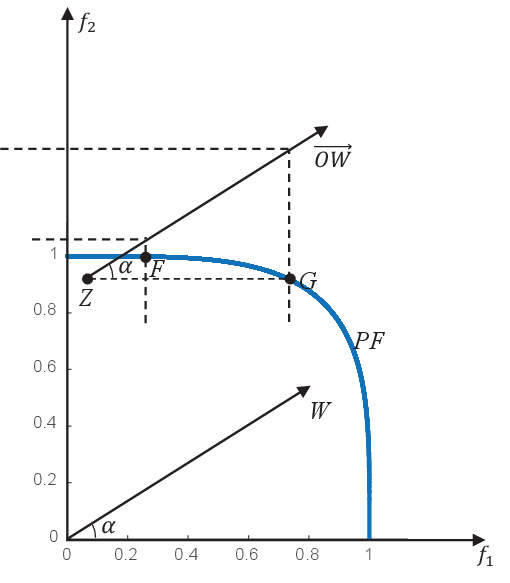}
  \caption{The objective function coordinate plane under the condition \(k_{OW} > k_{ZF}\)}
    \label{fig6.6}
\end{figure}
\begin{IEEEproof}
The objective function coordinate plane for this scenario is illustrated in Figure~\ref{fig6.6}. Clearly, by the geometric meaning of the M-TCH method,
\begin{equation*}
g^{mtch}(G|W,Z)=\frac{G_1-Z_1}{W_1}>\frac{F_1-Z_1}{W_1}=g^{mtch}(F|W,Z)
\end{equation*}

\end{IEEEproof}

Next, we discuss the scenario where \(k_{OW} \leq k_{ZF}\). We propose and prove the following theorem:
\begin{theorem}
\label{theorem4}
Given \(k_{OW} \leq k_{ZF}\), if
\begin{equation}
|k_G| \leq \tan\alpha
\end{equation}
then
\begin{equation}
g^{mtch}(F|W,Z) < g^{mtch}(G|W,Z)
\end{equation}
\end{theorem}

\begin{IEEEproof}
The objective function coordinate plane under this condition is illustrated in Figure~\ref{fig6.7}, where \(A\) is a point on the PF, and \(AZ \perp Of_1\). Clearly, by the geometric meaning of the M-TCH method,
\begin{equation*}
g^{mtch}(F|W,Z)=\frac{F_2-Z_2}{W_2}<\frac{|AZ|}{W_2}
\end{equation*}
\begin{equation*}
g^{mtch}(G|W,Z)=\frac{|GZ|}{W_1}
\end{equation*}

Therefore,
\begin{equation*}
\frac{g^{mtch}(G|W,Z)}{g^{mtch}(F|W,Z)} > \frac{W_2}{W_1} \frac{|GZ|}{|AZ|} =\frac{\tan \alpha} {|k_{AG}|} > \frac{\tan \alpha} {|k_{G}|} \geq 1
\end{equation*}

\begin{figure}[htbp]
  \centering
  \includegraphics[width=0.8\linewidth]{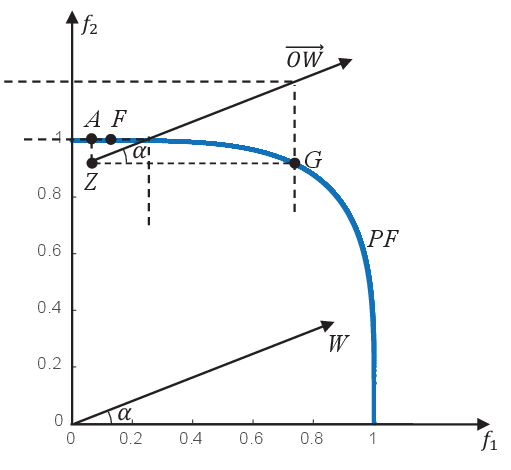}
  \caption{The objective function coordinate plane under the condition \(k_{OW} \leq k_{ZF}\)}
    \label{fig6.7}
\end{figure}
\end{IEEEproof}

Revisiting Theorem~\ref{theorem3} and Theorem~\ref{theorem4}, as long as the respective assumptions are satisfied, \(g^{mtch}(F|W,Z) < g^{mtch}(G|W,Z)\) holds true, meaning the fitness of individual \(G\) is worse than that of \(F\), implying the new individual \(G\) cannot be accepted, and the algorithm struggles to escape LO. As discussed in the Preliminaries chapter, the value of \(|k_G|\) is generally very small, thus as long as the value of \(\alpha\) is not equal to \(0\), the respective assumptions are easily satisfied. Therefore, the probability of the algorithm escaping LO is very small.

\section{Methodology}
\label{Methodology}

\begin{figure}[htbp]
\centering
\subfloat[]{
 \includegraphics[width=0.2\textwidth]{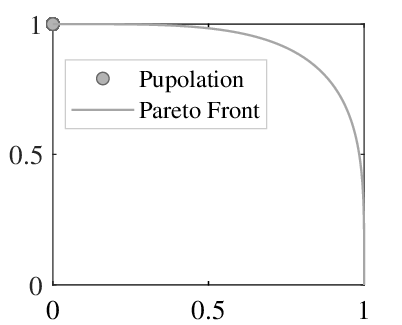}
\label{fg:intro1}}
\subfloat[]{
 \includegraphics[width=0.2\textwidth]{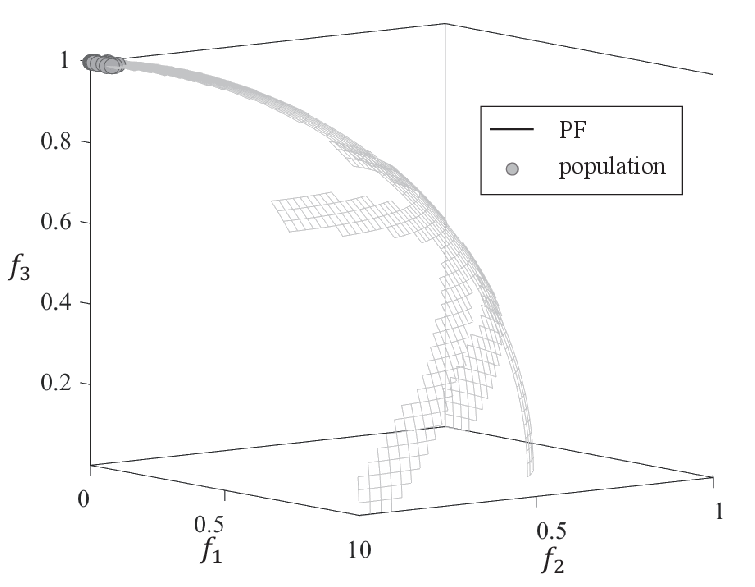}
\label{fg:intro2}}
\caption{Illustration of (a) MOEA/D on IMOP2 and (b) MOEADUR on IMOP7.}
\label{reference}
\end{figure}

As you can see in Figure~\ref{reference}, when MOEA/D-variants are applied to MOPs of non-convex PF, there is a high likelihood that they will degenerate into local search algorithms. This limits them to converging only to certain local optima on the PF, severely impacting the diversity of the population. Even recent algorithms (MOEADUR\cite{MOEADUR} published in 2022) developed in the last couple of years suffer from the same issue (see Figure~\ref{reference}).

\begin{figure}[htbp]
\centering
\subfloat[]{
 \includegraphics[width=0.2\textwidth]{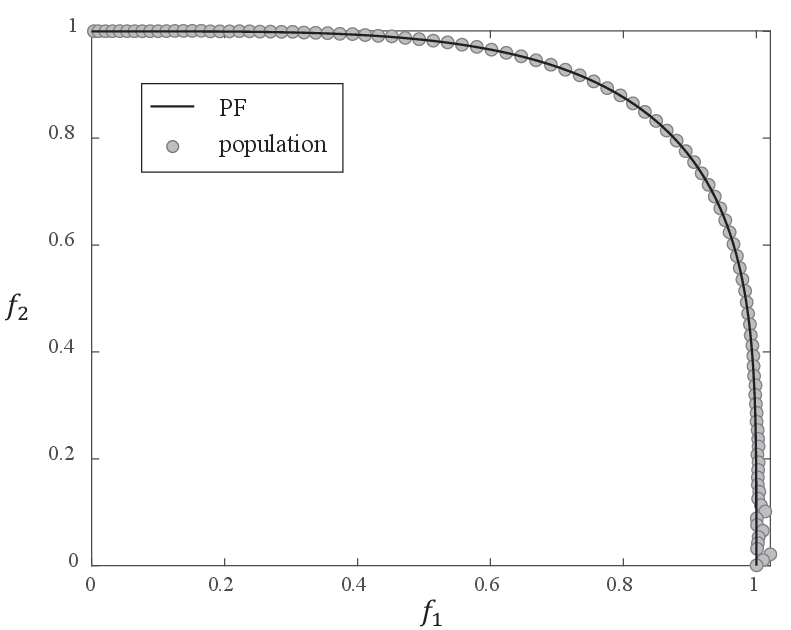}
\label{fg:moeadnoz}}
\subfloat[]{
 \includegraphics[width=0.2\textwidth]{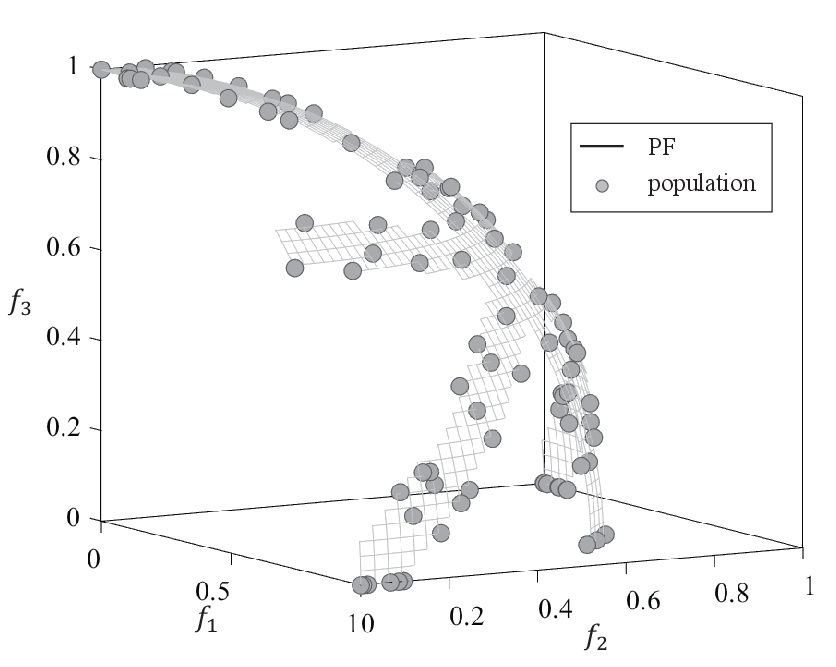}
\label{fg:durnoz}}
\caption{Illustration of (a) MOEA/D-normW  on IMOP2 and (b) MOEADUR-normW on IMOP7.}
\label{noz}
\end{figure}

Previous strategies have shown, as elaborated in Section \ref{sec:geometricProof}, that the conventional practice of choosing reference points might direct the MOEA/D variants towards local optima in scenarios involving concave challenges. The investigation reveals that the selection of reference points is crucial for these algorithms when addressing complex MOPs.

The detailed study presented in Section \ref{AblationStudy} indicates that substituting the reference point with the actual ideal point markedly improves the algorithm's population diversity and convergence, particularly in complex MOPs. Nevertheless, identifying the true ideal points in real-world MOPs is impractical, highlighting the necessity for a viable reference point selection method. Herein, we present a novel approach for determining the reference point during the algorithmic process, designated as NormW. This approach not only mitigates the issue of MOEA/D variants getting trapped in local optima but also maintains the algorithm's convergence speed on par with other methodologies. The ensuing section elaborates on our proposed technique.

As derived from the discussion on \textit{Review of Prior Approaches} \ref{sec:existingmethod}, current literature lacks a reference point selection method that effectively circumvents the local optima entrapment issue. The \textit{normW} method proposed herein addresses this concern adeptly. To integrate the \textit{normW} method into the MOEA/D variants algorithmic framework, one needs to substitute lines 8 and 13 in the MOEA/D pseudocode with the subsequent Algorithms \ref{alg:line8} and \ref{alg:line13}.

\begin{algorithm}[H]
\caption{Adjustment for Line 8}
\label{alg:line8}
\begin{algorithmic}[1]

\STATE Initialize the minimal reference point $Z_m = (z_i^{min})$.
\STATE Set the objective function's coordinate axis origin to $Z_0 = (0,...,0)$.
\STATE Establish the normal distribution's mean $\mu = maxFE/2$.
\STATE Set the normal distribution's standard deviation $\sigma = \mu/5$.

\end{algorithmic}
\end{algorithm}

\begin{algorithm}[H]
\caption{Modification for Line 13}
\label{alg:line13}
\begin{algorithmic}[1]
\STATE Update the minimal reference point $Z_m = \min(Z_m,y)$.
\STATE Determine the normW reference point $Z_w$.
\STATE Calculate the probability \( ( 1- \text{prob} ) \) for selecting $Z_w$.
\IF{\(\text{rand()} > \text{prob}\)}
\STATE Assign $Z = Z_w$.
\ELSE
    \IF{$\text{prob} < 1/2$}
        \STATE Set $Z = Z_0$.
    \ELSE 
    \STATE Use $Z = Z_m$.
\ENDIF
\ENDIF
\end{algorithmic}
\end{algorithm}

Subsequent sections will elucidate on the novel \(Z_w\), as detailed in Pseudocode \ref{alg:line13}, and the application of the Gaussian Distribution in Pseudocode \ref{alg:line8}.

\subsection{A Novel Preference Point $Z_w$}

The rationale for \(Z_w\) is detailed in Equation \ref{eq:Zwnormal}. Specifically, at each stage of evolution, the smallest objective values, denoted as \(Z_m\), are utilized to construct a quarter-circle (or for higher dimensions, a segment of a sphere) centered around the origin. The point where this shape intersects with the currently applied weight vector \(W_i\) yields the specific \(Z_w\) for that evolutionary iteration, labeled as \(Z_{W_i}\).

\begin{align}
\begin{aligned}
\overrightarrow{OZ}_{W_i} &=\left\| \overrightarrow{OZ_{m}}\right\| \times W_{i},\\
Z_{W_i} &= \overrightarrow{OZ}_{w_i}.
\end{aligned}
\label{eq:Zwnormal} 
\end{align}

\begin{figure}[htbp]
\centering
 \includegraphics[width=0.42\textwidth]{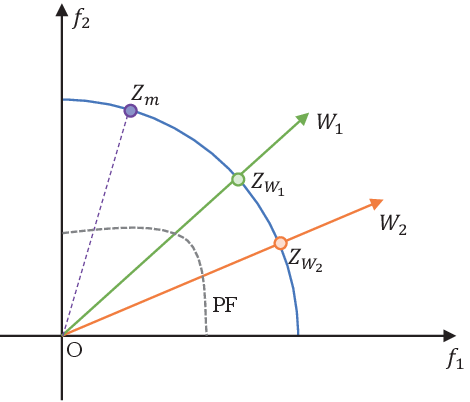}
 \caption{Illustration of $Z_w$ in relation to weight vectors.}
\label{fg:Zw}
\end{figure}

As illustrated in Figure \ref{fg:Zw}, \(Z_w\) demonstrates how it is harmonized with evenly spaced weight vectors, facilitating a wide-ranging set of solutions.

Such synchronization between \(Z_w\) and weight vectors plays a pivotal role in preserving solution diversity throughout the population's evolution, tackling the vital challenge of maintaining variation among generations.

\subsection{Gaussian Distribution}

A Gaussian distribution is defined for the variable \( Y \) with mean \( \mu \) and standard deviation \( \sigma \), where \( Y \) signifies the count of FE, also known as the generation number, as shown in Equation \ref{eq:normal_dis}. The choice of setting \( \sigma = \mu/5 \) stems from Equation \ref{eq:5sigma}, which posits that the likelihood \( p(Y=y) \) approximates 1 for \( y \) within the interval [0, \text{maxFE}], ensuring the algorithm cycles from 0 through \text{maxFE}.
\begin{align}
\begin{aligned}
Y&\sim \mathcal{N}(\mu,\sigma^2),\\
\mu &= \frac{\text{maxFE}}{2},\\
\sigma &= \frac{\mu}{5},
\end{aligned}
\label{eq:normal_dis} 
\end{align}

\begin{align}
\begin{aligned}
P&\left( 0\leq Y\leq \text{maxFE}\right) \\
&= p\left( -5\leq \frac{Y-\mu}{\sigma}\leq 5\right) \\
&= \Phi(5) - \Phi(-5) \\
&\approx 1,
\end{aligned}
\label{eq:5sigma} 
\end{align}
where \(\Phi(x)\) represents the cumulative distribution function (CDF) for the standard normal distribution.

The variable \( Y = \text{FE} \)'s CDF is articulated in Equation \ref{eq:CDF}. The probability that reference point \( Z \) is selected as either \( Z_0 \), \( Z_m \), or \( Z_w \) during the algorithm's evolution is depicted in Equation \ref{eq:pro}.
\begin{align}
    \begin{aligned}
        F(y; \mu, \sigma) = \frac{1}{\sigma\sqrt{2\pi}} \int_{-\infty}^{y} e^{-\frac{(y-\mu)^2}{2\sigma^2}} dy,
    \end{aligned}
    \label{eq:CDF}
\end{align}
\begin{align}
    \begin{aligned}
pro &= F(Y \leq y), \\
y &= \text{FE}, \\
p(Z = Z_w) &= 1 - pro, \\
p(Z = Z_{0}) &= pro, \text{ if } pro < \frac{1}{2}, \\
p(Z = Z_{m}) &= pro, \text{ if } pro \geq \frac{1}{2}.
\end{aligned}
    \label{eq:pro}
\end{align}

\begin{figure}[htbp]
\centering
 \includegraphics[width=0.5\textwidth]{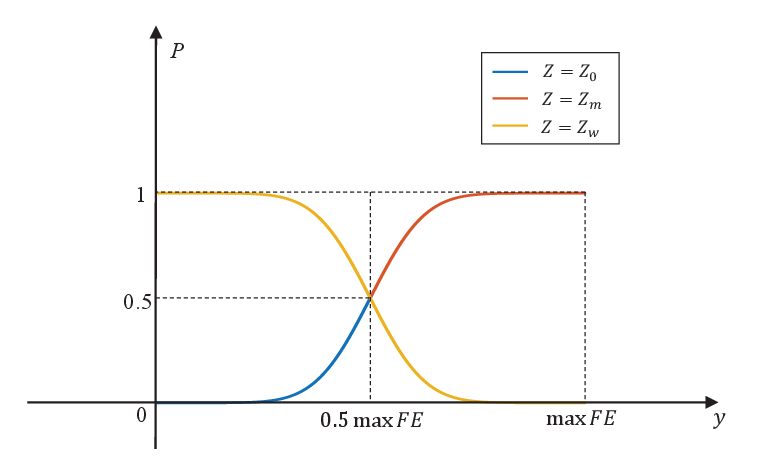}
 \caption{CDF showcasing the probabilities of selecting different reference points.}
\label{fg:normW}
\end{figure}
Figure \ref{fg:normW} illustrates the CDF of selecting various reference points, where the blue curve indicates the likelihood of opting for \( Z_0 \) within the first half of \text{maxFE}, the red curve represents the chance of selecting \( Z_m \) within the latter half, and the orange curve denotes the probability of choosing \( Z_w \).

The segmented choice mechanism adopted by the \textit{normW} method prioritizes $Z_w$ in the algorithm's initial phase to preserve population diversity and leans towards $Z_m$ in the concluding phase to assure convergence. Consequently, the \textit{normW} technique achieves a stable equilibrium between population diversity and convergence. Subsequently, a geometric analysis will corroborate that the \textit{normW} approach prevents MOEA/D-based algorithms from devolving into mere local search mechanisms when addressing complex MOPs.

\subsection{The Proof of Proposed Method}

Based on the RF selection strategy proposed in this paper, this chapter will analyze the capability of MOEA/D to escape local optima under the PBI and M-TCH methods, respectively. We still base our analysis on the model in Figure~\ref{fig6.3} and assess the possibility of the algorithm accepting a new individual \(G\).

First, the analysis results under the PBI method are presented as follows:
\begin{theorem}
\label{theorem5}
If
\begin{equation}
\tan\alpha \geq \theta^{-1}
\end{equation}
then
\begin{equation}
g^{pbi}(G|W,Z_W) < g^{pbi}(F|W,Z_W)
\end{equation}
\end{theorem}

\begin{IEEEproof}
The objective function coordinate plane under this scenario is illustrated in Figure~\ref{fig6.8}, where \(AF \perp OW\), \(CG \parallel OW\), and \(AF \parallel BG\). Clearly, by the geometric meaning of the PBI method,
\begin{equation*}
\begin{aligned}
&g^{pbi}(F|W,Z_W)-g^{pbi}(G|W,Z_W) \\&= (|AZ_W|+\theta|AF|)-(|BZ_W|+\theta|BG|) \\
&= (|AZ_W|-|BZ_W|) + \theta(|AF|-|BG|)\\
&\geq \theta|CF|-|AB|\\
&= |CG|(\theta \tan \angle{CGF}-1)\\
&> |CG|(\theta \tan \alpha-1) \geq 0
\end{aligned}
\end{equation*}

\begin{figure}[htbp]
  \centering
  \includegraphics[width=0.8\linewidth]{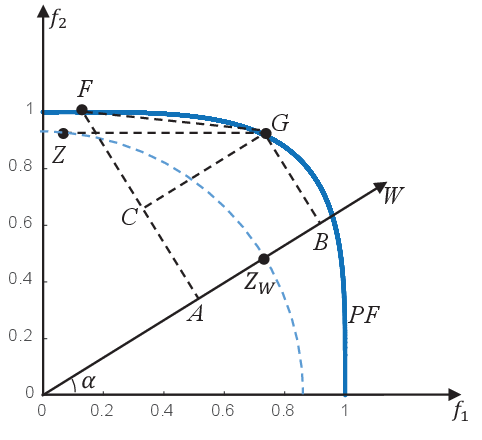}
  \caption{The objective function coordinate plane}
    \label{fig6.8}
\end{figure}
\end{IEEEproof}

Revisiting Theorem\ref{theorem5}, as long as the corresponding assumption is satisfied, \(g^{pbi}(G|W,Z) < g^{pbi}(F|W,Z)\) holds true, meaning the fitness of individual \(G\) is superior to that of \(F\), implying the new individual \(G\) can be accepted, allowing the algorithm to escape LO. Given that the default value of \(\theta\) is 5~\cite{MOEAD}, so for most weight vectors, the corresponding assumption is easy to be satisfied. In conclusion, the probability of the algorithms escaping LO is very high.

Next, the analysis results under the M-TCH method are presented as follows:
\begin{theorem}
\label{the6}
If
\begin{equation}
\tan\alpha \leq \frac{G_2}{G_1}
\end{equation}
then
\begin{equation}
g^{mtch}(G|W,Z_W) < g^{mtch}(F|W,Z_W)
\end{equation}
\end{theorem}

\begin{IEEEproof}
The objective function coordinate plane under this scenario is illustrated in Figure~\ref{fig6.9}. Clearly, by the geometric meaning of the M-TCH method,

\begin{equation*}
\begin{aligned}
g^{mtch}(G|W,Z_W)=&\frac{G_2-Z_{W2}}{W_2}\\
>&\frac{F_2-Z_{W2}}{W_2}\\
=&g^{mtch}(F|W,Z_W)
\end{aligned}
\end{equation*}

\begin{figure}[htbp]
  \centering
  \includegraphics[width=0.8\linewidth]{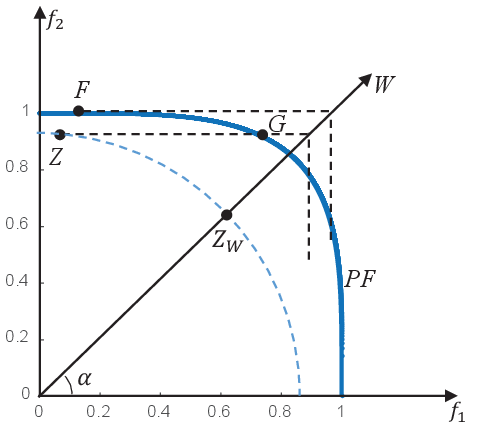}
  \caption{The objective function coordinate plane}
    \label{fig6.9}
\end{figure}

\end{IEEEproof}

Revisiting Theorem~\ref{the6}, as long as the corresponding assumption is satisfied, \(g^{mtch}(G|W,Z) < g^{mtch}(F|W,Z)\) holds true, meaning the fitness of individual \(G\) is superior to that of \(F\), implying the new individual \(G\) can be accepted, allowing the algorithm to escape LO. Notably, when \(Z_W\) is located below the intersection point of \(OG\) and the quarter circle, the assumption is guaranteed to be satisfied, hence the probability of the algorithm escaping LO is significant.


\section{Experiments}
\label{experiments}

In this section, we first demonstrate the significance of reference points on solution diversity in evolutionary algorithms through ablation studies. We selected algorithms published between 2014 and 2022, which are available on the PlatEMO platform (see Table 2). All these algorithms employ reference points during their execution. By setting the reference points directly to the true ideal point and comparing them with the original algorithms, the results indicates the method of choosing the reference points is the fundemental reason for MOEA/D to get trapped in local optima.



\subsection{Benchmark Test Problems}



For conventional algorithms that address multi-objective optimization problems (MOPs), the commonly adopted benchmark test problems are DTLZ and WFG. As shown in Table \ref{tab:IntroAlgrothm}, among the 14 algorithms we selected, 13 of them utilize either DTLZ or WFG. Since the main focus of this paper is the impact of the reference point on the diversity of the evolutionary algorithms, in addition to the DTLZ and WFG categories of test problems, we have included a series of  IMOP \cite{IMOP}  test problems, which can effectively distinguish the diversity performance of different algorithms.

As depicted in Table \ref{tab:testproblems} , we have selected a total of 16 benchmark test problems. Here, 'N' represents the population size, 'M' denotes the dimension of the objective vector, 'D' refers to the dimension of the decision vector, and 'MaxFE' indicates the maximum number of function evaluations. One evaluation process refers to obtaining the objective function value of a decision variable, then using a fitness function, which is mentioned in \ref{definition},  such as the decomposition method or the nondominated sorting method, also called PR \cite{NSGAII}, to determine the fitness value. This is done to decide whether to retain this decision variable or not.

\begin{table}[H]
    \centering
    \caption{The  experimental setting of the test problem}
    \renewcommand{\arraystretch}{1.4}{
    \small
    \begin{tabular}{ccccc}
    \hline
        Problem&$N$&$M$&$D$&$MaxFE$ \\ \hline

IMOP1 &100&2&10&20000 \\
IMOP2&100&2&10&20000 \\
IMOP3&100&2&10&20000 \\
IMOP4&100&3&10&20000 \\
IMOP5&100&3&10&20000 \\
IMOP6&100&3&10&20000 \\
IMOP7&100&3&10&20000 \\
IMOP8&100&3&10&20000 \\
WFG1&100&3&12&30000 \\
WFG2&100&3&12&30000 \\
WFG3&100&3&12&30000 \\
WFG4&100&3&12&30000 \\
DTLZ1&100&3&7&20000 \\
DTLZ2&100&3&12&20000 \\
DTLZ3&100&3&12&20000 \\
DTLZ4&100&3&12&20000 \\
\hline

    \end{tabular}}
\label{tab:testproblems}
\end{table}

\subsection{Selected Algorithms}






Over the past decade, algorithms based on the MOEA/D framework have emerged in abundance. Here, we have selected a total of 13 algorithms from the years 2014 to 2022, along with the MOEA/D algorithm, making it a total of 14, as shown in Table 2.

Out of the 14 algorithms we examined, 13 utilize the MOEA/D framework, but RVEAiGNG \cite{RVEAiGNG} does not. We included RVEAiGNG to demonstrate that when an algorithm employs a reference point, the method of selection can impact the diversity of its solutions. Among these 14 algorithms, 13 incorporate a reference point, whereas MOEADM2M does not. We picked MOEADM2M for its ability to maintain diversity within a population. Previously, it had been shown to increase diversity in comparison to MOEAD-based methods. In our study, we were able to achieve greater diversity than MOEADM2M by merely adjusting the reference point of the MOEA/D algorithm to the true ideal point. This finding emphasizes that the choice of reference point is a fundamental factor in the reduction of population diversity.

As indicated in Table 2, the fitness functions of these 14 algorithms encompass decomposition methods, PK, as well as new fitness functions proposed by certain algorithms, such as the Pareto adaptive scalarizing Methods introduced by MOEADPaS\cite{MOEADPas}.

Below, we will conduct a  ablation study using these 14 algorithms. And the parameter settings for these algorithms are all set to the default values on the PlatEMO platform\cite{platemo}.

\begin{table*}[htbp]
    \centering
    \caption{MOEA/D and Algorithms published from 2014 to 2022 that used the reference points }
    \renewcommand{\arraystretch}{1.4}
    \small
    \begin{tabular}{c|c|c|c|c}
    \toprule
        Algorithm & Publication Year & Based on MOEA/D & Fitness function&Test Problems \\ \hline
        MOEAD & 2007 & \ding{51} &  M-TCH &DTLZ  \\ 
        CMOEAD & 2014     & \ding{51}& PBI  & DTLZ, WFG\\ 
        MOEADAWA & 2014     & \ding{51} & TCH  & DTLZ\\ 
        MOEADM2M & 2014     & \ding{51} & PR & MOP \\ 
        MOEADMRDL & 2015     & \ding{51} & TCH & UF, WFG\\ 
        BCEMOEAD & 2016     & \ding{51} & M-TCH & DTLZ, WFG\\ 
        MOEADDU & 2016     & \ding{51} & M-TCH  & DTLZ, WFG\\ 
        MOEADPaS & 2016     & \ding{51} & PaS &WFG \\ 
        MOEADDCWV & 2019     & \ding{51} & M-TCH-N & WFG \\ 
        MOEADURAW & 2019     & \ding{51} & TCH  & WFG\\ 
        MOEADDYTS & 2020     & \ding{51} & TCH  & WFG, UF\\ 
        MOEADUR & 2022     & \ding{51} & TCH & DTLZ, WFG \\ 
        MOEADVOV & 2022     & \ding{51} & M-TCH  & DTLZ\\ 
        RVEAiGNG & 2022     & \ding{55} & Others  & DTLZ\\
\bottomrule
\end{tabular}
\label{tab:IntroAlgrothm}
\end{table*}

\subsection{Ablation Study}
\label{AblationStudy}

Previous research has been limited to enhancing the MOEA/D framework's algorithms for complex MOPs by improving the reference point, weight vectors, or decomposition methods. No studies have explicitly demonstrated the reasons why MOEA/D framework algorithms are sensitive to complex MOPs and tend to converge to local solutions on these issues. This paper, through the ablation study , identifies that the method of choosing the reference point is one of the fundamental reasons.


An Ablation Study\cite{ablation} in machine learning involves progressively removing parts of a model to understand their contribution to its performance. Here, we use the ablation study to compare the performance of the MOEA/D model before and after the removal of the reference point. 

 It's important to clarify that by 'removal' here, we mean the elimination of the method of choosing the reference point, using the true ideal point directly in MOEA/D-based algorithms. The advantage of this approach is that it can directly and effectively reflect the impact of the reference point selection method on the MOEA/D-based algorithms. In this paper, specifically, the most common method of selecting the reference point , which means min method, made these algorithms more likely to converge to local solutions and find it challenging to escape from them. And this directly results in an inability to obtain a well-distributed set of Pareto solutions, causing the algorithm to fail in maintaining population diversity. In the context of solving MOPs, this is a catastrophic characteristic. Thus it is one of the fundamental reasons why algorithms based on the MOEA/D framework struggle with complex MOPs.

Next, we will approach this from two categories of experimental results. The first category consists of 13 algorithms compared with their own versions that applied the true ideal point. The second category involves MOEADM2M compared with the MOEA/D algorithm that applied the true ideal point.

\begin{table}[htbp]
  \begin{minipage}{\linewidth}
    \centering
    \caption{The comparative results for the HV and IGD metrics between the selected algorithms and their modified counterparts evaluated over sixteen test problems and each run thirty times. }
    \renewcommand{\arraystretch}{1.45}
    \small
   \begin{tabular}{c|c|c|c}
  
    \toprule
       \multirow{2}{*}{ Algorithm }& HV &IGD  & \multirow{2}{*}{ Comparison } \\ &  \multicolumn{1}{c|}{($+/-/\approx$)}  &  \multicolumn{1}{c|}{($+/-/\approx$)}  \\ \hline
        MOEAD &      1  11 4 & 1  10  5        &   *-ideal    \\
        CMOEAD &       1  11 4 & 1 10  5     &   *-ideal    \\
        MOEADAWA &  0  7  9 &   0  8  8        &   *-ideal   \\
        MOEADM2M &    1  11  4  &  1  13  2        &   MOEAD-ideal   \\
        MOEADMRDL&    0  8  8  &  0  6  10         &  *-ideal  \\
        BCEMOEAD &    0  5  11  &  0  4  12      &   *-ideal \\
        MOEADDU &     0  11  5  &  0  10  6         &   *-ideal \\
        MOEADPaS &    0  7  9  &  0  5  11         &   *-ideal \\
        MOEADDCWV  &   1  6  9  &  1  6  9        &   *-ideal \\
        MOEADURAW   &  0  8  8  &  2  6  8      &   *-ideal \\
        MOEADDYTS   &  0  12  4  &  0  11  5          &   *-ideal \\
        MOEADUR &     0  9  7  &  1  6  9       &   *-ideal \\
        MOEADVOV &    2  8  6  &  1  8  7         &   *-ideal \\
        RVEAiGNG &    1  4  11  &  1  4  11         &   *-ideal \\

\bottomrule
\end{tabular}
    \label{ablationRe1}
  \end{minipage}%
  \end{table}

  \begin{table}
  \begin{minipage}{\linewidth}
    \centering
    \caption{The specific test problems where the original algorithms outperform the modified ones in terms of HV and IGD metrics.}
     \renewcommand{\arraystretch}{1.3}
     \small
    \begin{tabular}{c|c|c}
    \toprule
        Algorithm & \multicolumn{1}{c|}{HV  $+$}& \multicolumn{1}{c}{IGD  $+$}\\ \hline
        MOEAD &      WFG4   &  WFG4         \\
        CMOEAD &       WFG4   &  WFG4         \\
        MOEADAWA &   0        &   0   \\
        MOEADM2M &    IMOP1        &   IMOP1   \\
        MOEADMRDL&    0         & 0  \\
        BCEMOEAD &    0         & 0 \\
        MOEADDU &     0     &     0 \\
        MOEADPaS &    0         & 0 \\
        MOEADDCWV  &   WFG4     &  WFG4  \\
        MOEADURAW   &  0      &  IMOP5, IMOP8 \\
        MOEADDYTS   &    0      & 0  \\
        MOEADUR &     0        &  WFG4 \\
        MOEADVOV &    WFG3, DTLZ2         & IMOP5 \\
        RVEAiGNG &    WFG1        &  DTLZ2  \\

\bottomrule
\end{tabular}
    \label{ablationRE2}
  \end{minipage} 
\end{table}
\subsubsection{Algorithms for Self-Comparision}
\par

As you can see from Table \ref{ablationRe1}, all algorithms but MOEADM2M underwent an ablation study, comparing their original versions to those where the reference point was directly set to the true ideal point, referred to *-ideal (* standing for the specific algorithm). Across 16 benchmark test problems, results for HV and IGD metrics were derived. It's evident that in terms of the HV metric, which evaluates population diversity, all 13 algorithms showed significant enhancements. Additionally, the convergence metric, IGD, also observed improvements. For instance, the MOEADDYTS algorithm, released in 2020, when compared to its MOEADDYTS-ideal version, was outperformed in 12 of the 16 test problems for HV and in 11 problems for IGD. 

Table \ref{ablationRE2} illustrates the specific test problems from the set of 16, where the original algorithms outperform the modified ones in terms of HV and IGD metrics. For instance, for the MOEAD algorithm, it only outperforms the modified version on the WFG4 problem. This is attributed to the fact that the WFG4 problem\cite{WFG} is characterized by non-separability, multimodality, and non-convexity. As such, the number of evaluations required to converge to the PF is higher than other test problems. For simplicity, our study standardized the evaluation numbers for WFG problems at thirty thousand. This count may not suffice for some algorithms to reach a converged state on the WFG4 problem, leading to scenarios where the original algorithm outperforms the modified one. This also indicates that the convergence speed of algorithms using the true ideal point can be affected. 
As per the "No Free Lunch" theorem\cite{freelunch}, in the realm of algorithms, if there's a significant enhancement in population diversity on complicated MOPs, a slight compromise on convergence speed is acceptable. This reason applies to the WFG problems of CMOEAD, MOEADDCWV, and MOEADUR. Regarding the MOEADURAW algorithm\cite{MOEADURAW} for the IMOP5 and IMOP8 problems, the issues arise since both problems have discontinuous Pareto Fronts (PFs). The MOEADURW algorithm employs weight adaptation based on population sparsity and calculates the weight vector of the newly constructed subproblem using the reference point. The discontinuous PFs disrupt the algorithm's judgment on population sparsity. Additionally, when the method for selecting the reference point changes, the calculated weight vector deviates more from the original algorithm, affecting the convergence metric IGD for both problems.

An intriguing observation is that the final algorithm, RVEAiGNG\cite{RVEAiGNG}, is not based on MOEA/D, which employs reference vector adaptation for evolutionary optimization. Instead, it incorporates reference points within its algorithmic process. Insights from the ablation study of RVEAiGNG suggest that the selection method of reference points is pivotal for evolutionary algorithms to tackle complicated MOPs. An inappropriate approach can deteriorate the algorithm's global convergence capabilities.

Next, we will delve into a detailed comparison of the results between the MOEADM2M and MOEAD-ideal algorithms.

\subsubsection{MOEADM2M VS. MOEAD-ideal}


As previously mentioned, the introduction of MOEADM2M  aims to maintain population diversity, as evidenced by its superior HV metric when compared to other algorithms\cite{MOEADM2M}. In its originating paper, it was contrasted with an algorithm based on the MOEA/D framework (MOEADD) to highlight MOEADM2M's significant enhancement in population diversity. In this section, we continue to compare MOEADM2M with two algorithms based on the MOEA/D framework (MOEAD and MOEAD-ideal). The only modification in MOEA/D-ideal is setting the reference point to the true ideal point in the MOEA/D algorithm. As demonstrated in Tables \ref{ablationRe1} and \ref{ablationRE2}, out of the 16 benchmark test problems, only one test problems yielded a scenario where MOEADM2M outperformed MOEAD-ideal. Thus, from these comparative results, we can conclude that the selection method for the reference point is one of the fundamental reasons for the decline in population diversity.

\begin{figure*}[!t]
\centering
\subfloat[]{
 \includegraphics[width=0.65\textwidth]{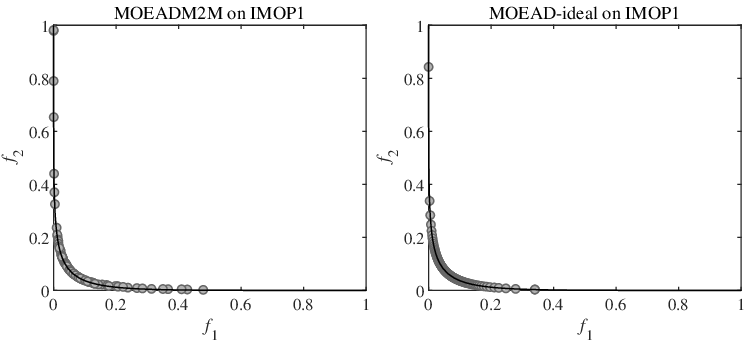}
\label{M2M_population}}
\hfil
\subfloat[]{
 \includegraphics[width=0.75\textwidth]{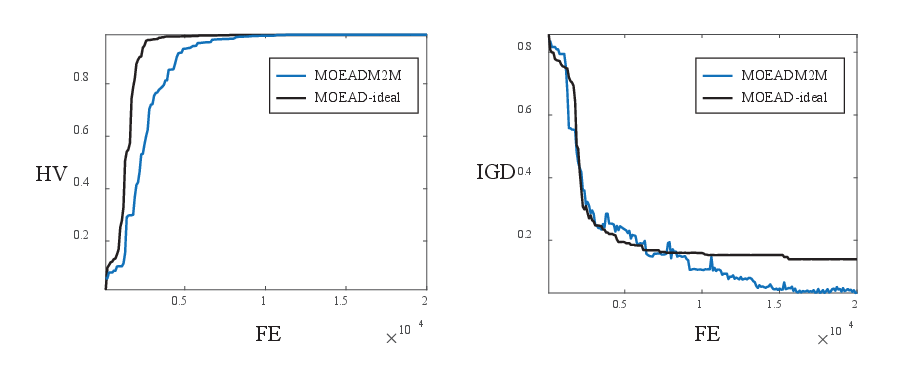}
\label{M2M_population}}
\caption{Graph representing the median of the value obtained from thirty runs of MOEADM2M and MOEAD-ideal (a) population (b) HV and IGD.}
\label{M2M_HV}
\end{figure*}

A more detailed result is provided in Table 5. The first column presents a comparison of the HV metrics for MOEADM2M, MOEAD, and MOEAD-ideal across 16 test problems, while the second column contrasts their IGD metrics. It's evident that MOEADM2M only outperforms MOEAD-ideal on the IMOP1 problem. For the IMOP1 problem, we ran both the MOEADM2M and MzOEAD-ideal algorithms thirty times each. The distribution graph corresponding to the median of the population is illustrated in Figure \ref{M2M_population}, where the black curve represents the PF. The metrics for HV and IGD, corresponding to their medians, are showcased in Figure \ref{M2M_HV}. As can be observed, the population distribution of MOEADM2M is more uniform, while MOEAD-ideal tends to concentrate around the center of the PF. This is because MOEADM2M solves multi-objective subproblems with each subproblem having a dedicated subpopulation that consistently receives computational effort during the search. It employs a strategy that sacrifices convergence speed to enhance population diversity. As seen in Figure \ref{M2M_HV}, MOEADM2M's HV converges after about 10,000 evaluations, while MOEAD-ideal's does so after roughly 3,000. MOEADM2M's IGD converges around 15,000 evaluations, compared to 5,000 for MOEAD-ideal. This clearly underscores MOEADM2M's emphasis on trading off convergence speed for population diversity. Additionally, due to the decomposition method limitations in MOEA/D\cite{IMOP}, it tends to concentrate populations in the central regions for convex problems, affecting overall distribution (HV) and convergence (IGD). Hence, we observe the phenomenon where MOEADM2M outperforms MOEAD-ideal on convex IMOP1 problem.
\begin{table*}[htbp]
\renewcommand{\arraystretch}{1.2}
\centering
\caption{The Mean value and standard deviation of the HV and IGD obtained by  MOEADM2M,MOEAD and MOEAD-ideal on sixteen test problems for thirty times. The best result on each problem is marked in grey.}
\resizebox{\textwidth}{!}{\begin{tabular}{cccc|ccc}
\toprule
Problem&MOEADM2M&MOEAD&MOEAD-ideal&MOEADM2M&MOEAD&MOEAD-ideal\\
\hline
\multirow{1}{*}{IMOP1}&\hl{9.8536e-1 (1.05e-3) $+$}&9.6738e-1 (2.02e-3) $-$&9.8476e-1 (7.81e-4)           &\hl{9.4028e-2 (2.47e-2) $+$}&3.6064e-1 (6.33e-3) $-$&1.5794e-1 (1.72e-2)\\
\hline                                                                                                    
\multirow{1}{*}{IMOP2}&1.7936e-1 (7.35e-3) $-$&9.0909e-2 (9.01e-8) $-$&\hl{2.1181e-1 (1.96e-2)}           &9.3790e-2 (2.22e-2) $-$&7.8497e-1 (2.30e-5) $-$&\hl{4.2917e-2 (2.37e-2)}\\
\hline                                                                                                    
\multirow{1}{*}{IMOP3}&5.8774e-1 (1.94e-2) $-$&3.4372e-1 (1.13e-1) $-$&\hl{6.6257e-1 (1.99e-2)}           &8.6407e-2 (7.04e-3) $\approx$&3.8187e-1 (1.26e-1) $-$&\hl{7.2917e-2 (3.01e-2)}\\
\hline                                                                                                    
\multirow{1}{*}{IMOP4}&3.9249e-1 (1.44e-2) $-$&1.9049e-1 (8.55e-2) $-$&\hl{4.2444e-1 (3.61e-4)}           &3.7295e-2 (8.14e-3) $-$&3.1988e-1 (1.75e-1) $-$&\hl{2.8990e-2 (5.51e-4)}\\
\hline                                                                                                    
\multirow{1}{*}{IMOP5}&3.3339e-1 (6.46e-3) $-$&1.4669e-1 (3.10e-4) $-$&\hl{3.5220e-1 (5.79e-3)}           &7.1619e-2 (8.13e-3) $-$&5.2979e-1 (3.70e-4) $-$&\hl{6.0556e-2 (1.60e-2)}\\
\hline                                                                                                    
\multirow{1}{*}{IMOP6}&2.8112e-1 (4.55e-2) $-$&3.2191e-1 (2.12e-1) $\approx$&\hl{5.1433e-1 (1.14e-3)}     &2.3732e-1 (6.65e-2) $-$&2.6791e-1 (2.42e-1) $\approx$&\hl{4.9885e-2 (1.79e-3)}\\
\hline                                                                                                    
\multirow{1}{*}{IMOP7}&2.1245e-1 (8.55e-2) $-$&9.0909e-2 (1.24e-7) $-$&\hl{3.3377e-1 (1.81e-1)}           &4.5928e-1 (2.57e-1) $-$&9.3878e-1 (1.19e-4) $-$&\hl{3.8275e-1 (3.73e-1)}\\
\hline                                                                                                    
\multirow{1}{*}{IMOP8}&2.6326e-1 (3.66e-2) $-$&6.9956e-2 (1.38e-3) $-$&\hl{6.0921e-1 (6.16e-3)}           &3.4630e-1 (3.86e-2) $-$&1.0629e+0 (1.81e-3) $-$&\hl{1.2376e-1 (4.07e-3)}\\
\hline
\multirow{1}{*}{WFG1}&3.8184e-1 (2.20e-2) $-$&8.4449e-1 (3.17e-2) $-$&\hl{8.6873e-1 (1.76e-2)}           &1.3213e+0 (7.13e-2) $-$&3.1604e-1 (3.32e-2) $-$&\hl{2.7362e-1 (1.55e-2)}\\
\hline                                                                                                   
\multirow{1}{*}{WFG2}&8.7978e-1 (8.89e-3) $-$&8.9371e-1 (9.45e-3) $-$&\hl{8.9772e-1 (1.17e-2)}           &2.8766e-1 (1.39e-2) $-$&2.5257e-1 (7.74e-3) $\approx$&\hl{2.4814e-1 (2.70e-2)}\\
\hline                                                                                                   
\multirow{1}{*}{WFG3}&2.9707e-1 (1.54e-2) $-$&3.5117e-1 (1.18e-2) $-$&\hl{3.5710e-1 (5.49e-3)}           &3.0264e-1 (2.55e-2) $-$&1.7419e-1 (2.51e-2) $-$&\hl{1.6119e-1 (9.31e-3)}\\
\hline                                                                                                   
\multirow{1}{*}{WFG4}&4.8041e-1 (7.99e-3) $-$&\hl{5.2912e-1 (3.55e-3) $+$}&5.1832e-1 (5.37e-3)           &3.5393e-1 (1.63e-2) $-$&\hl{2.6352e-1 (5.46e-3) $+$}&2.9806e-1 (1.35e-2)\\
\hline
\multirow{1}{*}{DTLZ1}&3.3523e-2 (1.25e-1) $-$&\hl{8.2106e-1 (4.68e-2) $\approx$}&7.9002e-1 (1.67e-1)     &3.7184e+0 (2.37e+0) $-$&\hl{2.4996e-2 (1.27e-2) $\approx$}&4.4656e-2 (9.74e-2)\\
\hline                                                                                                    
\multirow{1}{*}{DTLZ2}&3.5897e-1 (1.31e-2) $-$&5.5897e-1 (1.25e-4) $\approx$&\hl{5.5898e-1 (1.47e-4)}     &1.5196e-1 (6.08e-3) $-$&5.4485e-2 (8.42e-6) $\approx$&\hl{5.4483e-2 (5.16e-6)}\\
\hline                                                                                                    
\multirow{1}{*}{DTLZ3}&0.0000e+0 (0.00e+0) $-$&\hl{2.9428e-1 (2.16e-1) $\approx$}&2.2762e-1 (2.27e-1)     &5.3264e+1 (1.72e+1) $-$&\hl{6.6947e-1 (9.61e-1) $\approx$}&7.4065e-1 (8.72e-1)\\
\hline                                                                                                    
\multirow{1}{*}{DTLZ4}&4.5854e-1 (1.17e-2) $\approx$&3.5373e-1 (1.87e-1) $-$&\hl{4.6244e-1 (1.12e-1)}     &\hl{1.0504e-1 (4.72e-3) $\approx$}&4.7110e-1 (3.60e-1) $-$&2.6630e-1 (2.46e-1)\\
\hline
\multicolumn{1}{c}{HV$+/-/\approx$}&1/14/1&1/11/4  &  &1/13/2&1/10/5&\multicolumn{1}{c}{IGD($+/-/\approx$)}\\
\bottomrule
\end{tabular}}
\label{m2m-ideal}
\end{table*}



\subsection{Experiments for Proposed Method}
\label{sec:expProposed}

From our discussion in the ablation study section\ref{AblationStudy}, it's clear that the commonly used min method can lead MOEA/D-based algorithms to degrade from a global search nature to a local search, especially when addressing complex MOPs like concave problems. To address this, we utilize the normW method introduced in the methodology chapter\ref{Methodology}, demonstrating through empirical evidence that this method can effectively resolve the issue. The experiments in this section are twofold:

First, we compare MOEA/D-based algorithms employing the min method against those using the normW method. From this comparison, it's evident that the normW method, relative to the min method, exhibits improvements in both diversity and convergence of the population when solving complex MOPs.

Secondly,we compare the MOEAD algorithm that employs the DRP method, a latest approach for reference point selection\cite{otherMethod}, with the MOEAD algorithm that utilizes the normW method. The aim here is to emphasize that our proposed method outperforms the latest method, demonstrating its potential to enhance both population diversity and convergence in MOEA/D-based algorithms.

\subsubsection{Comparison with the traditional Method}
In all the algorithms we've chosen, the most prevalent method, min, has been uniformly used for selecting the reference point. In this subsection, our primary focus is to compare the effects on population diversity and convergence when the chosen algorithms employ either the min or normW selection methods.
In this stage, the benchmark remains consistent with 16 test problems, and all experimental settings are kept unchanged, as shown in Table 1. There were originally 14 selected algorithms, as depicted in Table 2. However, since the normW method employs uniformly distributed weight vectors, we excluded algorithms related to adaptive weight vectors during the experimentation. This includes MOEADAWA, MOEADURAW, MOEADDCWV, and RVEAiGNG. For such algorithms, our recommended improvement step is to first use the generated uniformly distributed weight vectors combined with the normW method to determine the reference point before acquiring the adaptive weight vectors. For instance, the algorithm MOEADUR, which relates to adaptive weight vectors, yielded results as shown in Table \ref{normWMIN} after the above modification. We also removed certain algorithms that use two types of reference points in their procedures, which include MOEADDU and MOEADPaS.
As illustrated in Table 6, when the remaining algorithms switched their reference point selection method to normW, they are denoted as *-normW (where * stands for the original algorithm), and the results were then compared with those of the original algorithms. Notably, regardless of whether we consider the HV or IGD metrics, all seven selected algorithms exhibited varying degrees of improvement across the 16 test problems. For instance, MOEADM2M trailed behind MOEAD-normW in the HV metric on 13 test problems and in the IGD metric on 11 test problems. As discussed in the ablation study section \ref{AblationStudy}, MOEADM2M was primarily designed to retain population diversity. Our proposed normW method, applied only to the most basic MOEA/D algorithm, can achieve commendable results even when compared to MOEADM2M. This underscores the efficacy of our approach and highlights the pivotal role the choice of reference point plays in preserving population diversity for algorithms based on the MOEA/D framework.

\begin{table}
    \centering
    \caption{The comparison results of the HV and IGD obtained by selected algorithms  on sixteen test problems for thirty times.}
    \renewcommand{\arraystretch}{1.5}
    \small
   \begin{tabular}{c|c|c|c}
  
    \toprule
       \multirow{2}{*}{ Algorithm }& HV &IGD  & \multirow{2}{*}{ Comparison } \\ &  \multicolumn{1}{c|}{($+/-/\approx$)}  &  \multicolumn{1}{c|}{($+/-/\approx$)}  \\ \hline
        MOEAD &      1  11 4 & 1  10  5        &   *-normW    \\
        CMOEAD &       3  7 6 & 3 7  6     &   *-normW    \\
        MOEADM2M &    2  13  1  &  2  11  3        &   MOEAD-normW   \\
        MOEADMRDL&    4  10  2  & 4  9  3        &  *-normW   \\
        BCEMOEAD &    0  3  13  &  0  3  13      &   *-normW \\
        MOEADUR &     1  9  6  &  0  5  11       &   *-normW \\
        MOEADVOV &    3  9  4  &  2  8  6         &  *-normW  \\ 
\bottomrule
\end{tabular}
    \label{normWMIN}
\end{table}

\subsubsection{Comparison with the latest Method }


To the best of our knowledge regarding research on the selection of reference points, the DRP method\cite{otherMethod}  is the most widely applied and represents the most recent findings. In this section, we apply both the DRP method and our proposed normW method as selection strategies of reference points to the MOEAD algorithm and compare their performances across sixteen test problems. The results are presented in Table \ref{DRP_Hv}. When evaluating convergence using the IGD metric, the normW method outperforms the DRP method in 11 of the test problems. In terms of population diversity, as measured by the HV metric, the normW method surpasses the DRP method in 10 test problems.

\begin{table*}[htbp]
\renewcommand{\arraystretch}{1.2}
\centering
\caption{The Mean value and standard deviation of the HV and IGD obtained by MOEAD-DRP and MOEAD-normW
on sixteen test problems for thirty times. The best result on each problem is marked in grey.}
\resizebox{\textwidth}{!}{
\small
\begin{tabular}{ccccc|cc}
\toprule
Problem&$M$&$D$&MOEAD-DRP&MOEAD-normW&MOEAD-DRP&MOEAD-normW\\
\hline
\multirow{1}{*}{DTLZ1}&3&7&7.7932e-1 (1.18e-1) $-$&\hl{8.3523e-1 (4.43e-3)}                               &4.4239e-2 (4.76e-2) $-$&\hl{2.2044e-2 (1.20e-3)}\\
\hline                           
\multirow{1}{*}{DTLZ2}&3&12&5.4059e-1 (3.24e-3) $-$&\hl{5.5914e-1 (2.22e-4)}                          &8.0873e-2 (3.34e-3) $-$&\hl{5.5589e-2 (4.41e-4)}\\
\hline                           
\multirow{1}{*}{DTLZ3}&3&12&4.4552e-2 (1.21e-1) $-$&\hl{2.9755e-1 (2.25e-1)}                          &2.8763e+0 (1.98e+0) $-$&\hl{5.9333e-1 (7.50e-1)}\\
\hline                           
\multirow{1}{*}{DTLZ4}&3&12&3.4309e-1 (1.55e-1) $-$&\hl{4.1796e-1 (1.53e-1)}                          &5.0060e-1 (3.00e-1) $-$&\hl{3.5598e-1 (3.07e-1)}\\
\hline                           
\multirow{1}{*}{IMOP1}&2&10&9.8464e-1 (9.61e-4) $-$&\hl{9.8556e-1 (5.33e-4)}                          &\hl{9.0767e-2 (1.41e-2) $+$}&1.3129e-1 (2.26e-2)\\
\hline                           
\multirow{1}{*}{IMOP2}&2&10&\hl{2.2790e-1 (8.78e-3) $\approx$}&2.1274e-1 (1.96e-2)                           &\hl{1.7861e-2 (1.60e-2) $\approx$}&3.4922e-2 (2.99e-2)\\
\hline                           
\multirow{1}{*}{IMOP3}&2&10&\hl{6.4245e-1 (1.48e-2) $\approx$}&6.3249e-1 (1.62e-2)                           &\hl{4.6526e-2 (3.74e-2) $\approx$}&6.7555e-2 (3.33e-2)\\
\hline                           
\multirow{1}{*}{IMOP4}&3&10&4.2223e-1 (1.46e-3) $-$&\hl{4.2463e-1 (2.34e-4)}                           &2.9924e-2 (1.85e-3) $-$&\hl{2.9083e-2 (4.48e-4)}\\
\hline                           
\multirow{1}{*}{IMOP5}&3&10&\hl{3.3526e-1 (7.38e-2) $+$}&2.9298e-1 (9.98e-2)                           &\hl{1.5807e-1 (1.47e-1) $\approx$}&2.3815e-1 (1.99e-1)\\
\hline                           
\multirow{1}{*}{IMOP6}&3&10&5.0461e-1 (1.87e-3) $-$&\hl{5.1642e-1 (7.53e-4)}                           &5.9413e-2 (2.75e-3) $-$&\hl{4.7894e-2 (1.09e-3)}\\
\hline                           
\multirow{1}{*}{IMOP7}&3&10&2.7159e-1 (1.94e-1) $-$&\hl{4.1268e-1 (1.78e-1)}                           &5.1689e-1 (4.05e-1) $-$&\hl{2.5034e-1 (3.56e-1)}\\
\hline                           
\multirow{1}{*}{IMOP8}&3&10&4.3909e-1 (1.33e-2) $-$&\hl{4.8077e-1 (2.81e-3)}                           &1.9449e-1 (1.31e-2) $-$&\hl{1.4808e-1 (3.31e-3)}\\
\hline                           
\multirow{1}{*}{WFG1}&3&12&9.0791e-1 (8.57e-3) $-$&\hl{9.2046e-1 (3.04e-3)}                        &2.5052e-1 (1.78e-2) $-$&\hl{2.0542e-1 (9.69e-3)}\\
\hline                           
\multirow{1}{*}{WFG2}&3&12&\hl{8.9413e-1 (1.14e-2) $+$}&8.7553e-1 (1.19e-2)                          &2.7626e-1 (1.45e-2) $-$&\hl{2.5298e-1 (8.97e-3)}\\
\hline                           
\multirow{1}{*}{WFG3}&3&12&\hl{3.7406e-1 (2.87e-3) $+$}&3.7266e-1 (1.29e-3)                          &\hl{1.3076e-1 (5.98e-3) $+$}&1.3643e-1 (5.71e-4)\\
\hline                           
\multirow{1}{*}{WFG4}&3&12&\hl{5.2340e-1 (6.37e-3) $+$}&5.1069e-1 (1.02e-2)                          &3.2030e-1 (1.47e-2) $-$ &\hl{3.0059e-1 (1.17e-2)}\\
\hline
\multicolumn{1}{c}{HV$+/-/\approx$}&  &  &4/10/2   & &2/11/3&\multicolumn{1}{c}{IGD($+/-/\approx$)}\\
\bottomrule
\end{tabular}}
\label{DRP_Hv}
\end{table*}

\begin{figure}[htbp]
\centering
\subfloat[]{\includegraphics[width=0.48\textwidth]{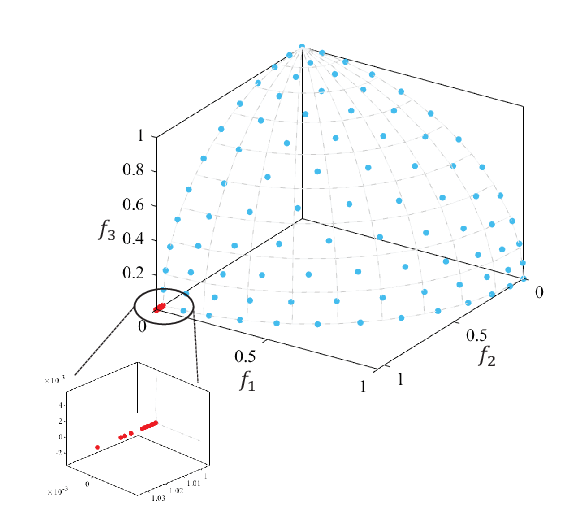}%
\label{DRP_DTLZ4}}
\hfil
\subfloat[]{\includegraphics[width=0.52\textwidth]{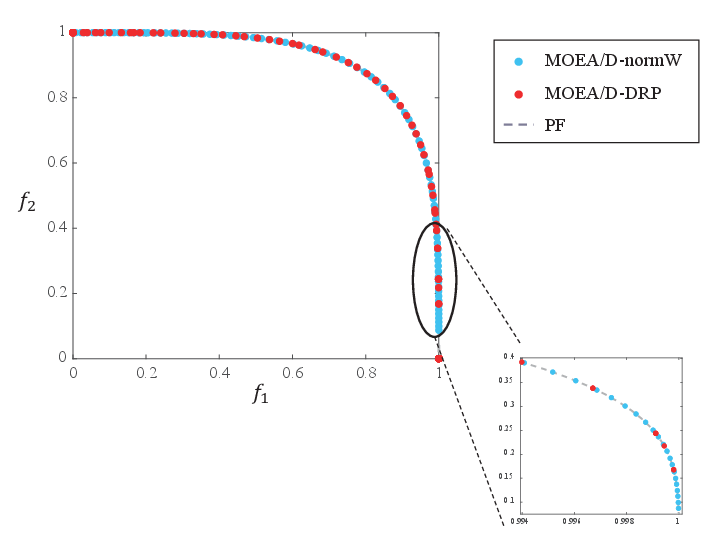}%
\label{DRP_IMOP2}}
\caption{Graph representing the median of the population obtained from thirty runs of the MOEAD-DRP and MOEA/D-normW (a) DTLZ4. (b) IMOP2.}
\label{DRP_COMPLEX}
\end{figure}

Considering our primary focus is on complex MOPs, we have selected IMOP2 and DTLZ4—both highly concave problems—for a comparative study of the populations. As seen in Figure\ref{DRP_DTLZ4}, for DTLZ4, the DRP method still frequently converges to local solutions. And in contrast, the normW method converges to a set of evenly distributed solutions. Similarly, as depicted in Figure \ref{DRP_IMOP2} for IMOP2, the population obtained using the DRP method is not as uniformly distributed as that of the normW method. This demonstrates the capability of the normW method to handle complex MOPs effectively.

\begin{figure*}[htbp]
\centering
\subfloat[]{\includegraphics[width=2.3in]{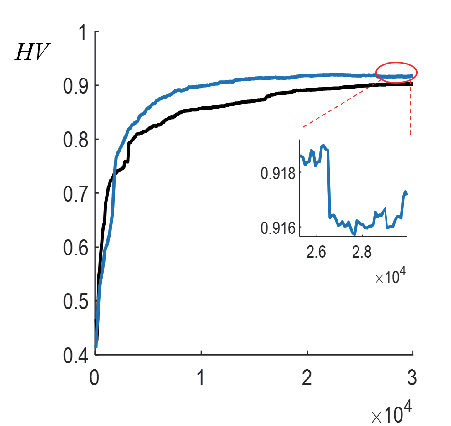}%
\label{DRP_WFG2}}
\hfil
\subfloat[]{\includegraphics[width=2in]{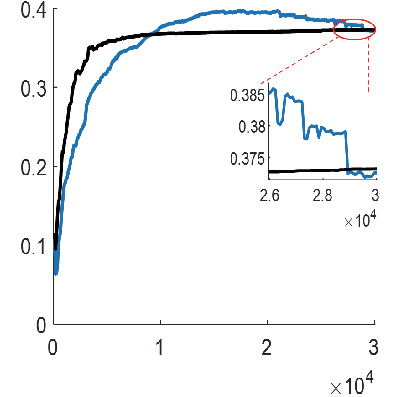}%
\label{DRP_WFG3}}
\hfil
\subfloat[]{\includegraphics[width=2.2in]{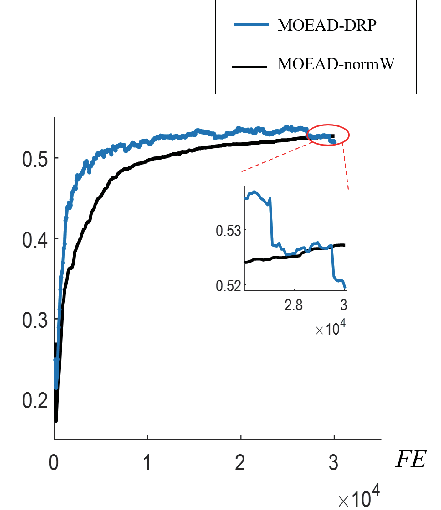}%
\label{DRP_WFG4}}
\caption{This graph corresponding to the median HV from 30 runs for MOEAD-DRP and MOEAD-normW on the (a) WFG2 (b)WFG3 and (c) WFG4 problems .}
\label{DRP_WFG}
\end{figure*}

Regarding the WFG problems, as can be seen from Table \ref{DRP_Hv}, the DRP method outperforms the normW method in terms of average HV on WFG2, WFG3, and WFG4. Figure \ref{DRP_WFG} depicts the HV obtained by the algorithms over 30,000 evaluation iterations. As clearly illustrated in Figure \ref{DRP_WFG}, the most prominent difference between the two is that, after convergence, the MOEAD-normW algorithm tends to stabilize and shows a gradual increase. In contrast, the MOEAD-DRP algorithm not only exhibits significant fluctuations after convergence but also occasionally demonstrates a declining trend, as evidenced in Figures \ref{DRP_WFG3} and \ref{DRP_WFG4}. This suggests that, although MOEAD-DRP might surpass MOEAD-normW in terms of average values, MOEAD-normW generally yields a more stable population in the latter stages of the algorithm. Due to the linear relationship between the number of evaluations and the reference point formula in the DRP method \cite{otherMethod}, it continues to fluctuate and decline even after convergence. We can infer from this that if the number of evaluations exceeds 30,000, say 50,000, the HV of MOEAD-normW would gradually surpass that of the MOEAD-DRP.
 For the WFG3 problem, neither method converges fully to the PF within 50,000 evaluations, as depicted in Figure 8, which makes it challenging to determine which one is superior post-convergence. However, the DRP method appears to have the edge over the normW method before convergence. In summary, across the 16 test problems, the normW method significantly outperforms the DRP method, especially when the WFG problems converge. Thus, we have grounds to believe that substituting the reference point selection method in most MOEA/D-based algorithms with the normW method can enhance their performance to varying extents. This reaffirms the pivotal role of reference point selection in MOEA/D-based algorithms addressing complex MOPs.



\section{Conclusion and future works}
\label{conclusion}



Based on the analyses and experiments conducted, we can confidently assert that the traditional method of selecting the reference point is one of the fundamental reasons for the performance degradation of MOEA/D-based algorithms when addressing complex MOPs. To tackle this issue, we introduced the \textit{normW} method. We utilized a normal distribution to determine the probability of selecting different reference points at various stages. Innovatively, we designed a reference point, termed \( Zw \), that allows the algorithm to escape local solutions. By leveraging the strengths of both approaches, we achieved significant improvements in population diversity and convergence compared to previous methods.

For future research, we aim to explore the influence of the reference point on algorithmic performance, not restricting ourselves solely to MOEA/D-based algorithms. In this paper, although we selected only one multi-objective evolutionary algorithm not based on the MOEA/D framework (namely RVEAiGNG), the experimental results for the benchmark test problems improved in both the ablation study and when applying the ANormW method. This indicates that the reference point plays an indispensable role in the global search capability of evolutionary algorithms.


\bibliographystyle{IEEEtran}
\bibliography{references}

\end{document}